\documentclass{article}
\usepackage[nonatbib,final]{neurips_data_2022}

\usepackage[utf8]{inputenc} %
\usepackage[T1]{fontenc}    %
\usepackage[square,comma,numbers]{natbib}

\usepackage{amsmath,amsfonts,bm}

\def\eqref#1{equation~\ref{#1}}

\def\1{\bm{1}}

\DeclareMathAlphabet{\mathsfit}{\encodingdefault}{\sfdefault}{m}{sl}
\SetMathAlphabet{\mathsfit}{bold}{\encodingdefault}{\sfdefault}{bx}{n}

\def\gA{{\mathcal{A}}}

\def\gC{{\mathcal{C}}}

\def\gG{{\mathcal{G}}}

\def\gS{{\mathcal{S}}}

\def\gU{{\mathcal{U}}}
\def\gV{{\mathcal{V}}}

\newcommand{\R}{\mathbb{R}}

\usepackage{color,xcolor}
\usepackage{epsfig}
\usepackage{graphicx}

\usepackage{adjustbox}
\usepackage{array}
\usepackage{booktabs}
\usepackage{colortbl}
\usepackage{wrapfig}
\usepackage{hhline}
\usepackage{multirow}
\usepackage{subcaption} %
\usepackage{wrapfig}
\usepackage{floatflt}

\usepackage{amsmath,amsfonts,amssymb}
\usepackage{mathtools}  %
\usepackage{bm}
\usepackage{nicefrac}
\usepackage{microtype}

\usepackage{changepage}
\usepackage{extramarks}
\usepackage{fancyhdr}
\usepackage{lastpage}
\usepackage{setspace}
\usepackage{soul}
\usepackage{xspace}

\usepackage[pagebackref=true,breaklinks=true,colorlinks,citecolor=gray]{hyperref}

\usepackage{url}

\usepackage{enumerate}
\usepackage{todonotes} %
\usepackage{enumitem}  %

\usepackage{titlesec}

\usepackage{makecell}

\usepackage{pifont} %

\usepackage{algorithm,algpseudocode}

\usepackage{amsthm}
\usepackage{framed}
\definecolor{shadecolor}{gray}{0.95}

\usepackage{bbding}
\newcolumntype{L}[1]{>{\raggedright\let\newline\\\arraybackslash\hspace{0pt}}m{#1}}
\newcolumntype{C}[1]{>{\centering\let\newline\\\arraybackslash\hspace{0pt}}m{#1}}
\newcolumntype{R}[1]{>{\raggedleft\let\newline\\\arraybackslash\hspace{0pt}}m{#1}}

\newcommand{\sect}[1]{Section~\ref{#1}}

\newcommand{\fig}[1]{Fig.~\ref{#1}}
\newcommand{\tbl}[1]{Table~\ref{#1}}

\newcommand{\ignore}[1]{}

\makeatletter
\DeclareRobustCommand\onedot{\futurelet\@let@token\@onedot}
\def\@onedot{\ifx\@let@token.\else.\null\fi\xspace}

\def\eg{e.g\onedot} 
\def\ie{i.e\onedot} 
 
\def\etc{etc\onedot}

\def\etal{et al\onedot}

\makeatother

\definecolor{MyDarkBlue}{rgb}{0,0.08,1}
\definecolor{MyDarkGreen}{rgb}{0.02,0.6,0.02}
\definecolor{MyDarkRed}{rgb}{0.8,0.02,0.02}
\definecolor{MyDarkOrange}{rgb}{0.40,0.2,0.02}
\definecolor{MyPurple}{RGB}{111,0,255}
\definecolor{MyRed}{rgb}{1.0,0.0,0.0}
\definecolor{MyGold}{rgb}{0.75,0.6,0.12}
\definecolor{MyDarkgray}{rgb}{0.66, 0.66, 0.66}

\newcommand{\tbd}[1]{\colorbox{MyRed}{\textcolor{white}{??}}\xspace}

\usepackage{pifont} %

\newcommand{\benchmark}{HandMeThat\xspace}
\newcommand{\behavior}{BEHAVIOR-100\xspace}

\newcommand{\xhdr}[1]{\noindent\textbf{#1}}

\newcommand{\mycell}[1]{\begin{tabular}[t]{@{}l@{}l}#1\end{tabular}}
\newcommand{\mycellc}[1]{\begin{tabular}[t]{@{}c@{}l}#1\end{tabular}}

\newcommand{\revise}[1]{#1}
\newcommand{\camera}[1]{#1}
 
\title{\benchmark: Human-Robot Communication\\ in Physical and Social Environments}
\author{
Yanming Wan\thanks{indicates equal contribution. Correspondence to: {\tt jiayuanm@mit.edu}.\\Project website: \url{http://handmethat.csail.mit.edu/}}\\
IIIS, Tsinghua University
\And
Jiayuan Mao$^*$\\
MIT CSAIL
\And
Joshua B. Tenenbaum\\
MIT BCS, CBMM, CSAIL
}

\begin{document}

\maketitle

\begin{abstract}
We introduce \benchmark, a benchmark for a holistic evaluation of instruction understanding and following in physical and social %
\revise{environments}. While previous datasets primarily focused on language grounding and planning, \benchmark considers the resolution of human instructions with {\it ambiguities} based on the physical (object states and relations) and social (human actions and goals) %
\revise{information}. \benchmark contains 10,000 episodes of human-robot interactions. In each episode, the robot first observes a trajectory of human actions towards her internal goal. Next, the robot receives a human instruction and should take actions to accomplish the subgoal set through the instruction.
In this paper, we present a textual interface for our benchmark, where the robot interacts with a virtual environment through textual commands. We evaluate several baseline models on \benchmark, and show that both offline and online reinforcement learning algorithms perform poorly on \benchmark, suggesting significant room for future work on physical and social human-robot communications and interactions.
\end{abstract}
\section{Introduction}

\begin{figure}[tp]
  \centering\small
  \includegraphics[width=\textwidth]{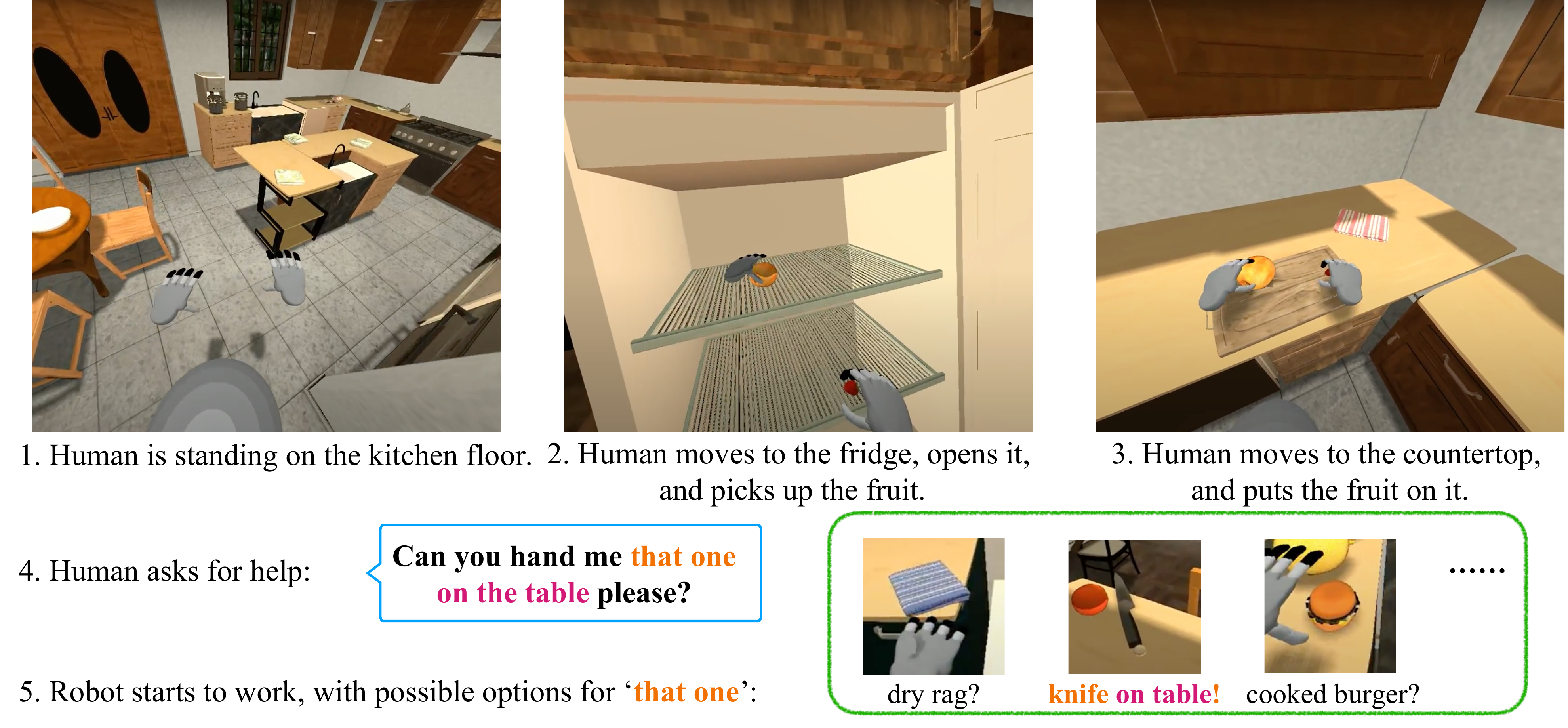}
  \caption{An example HandMeThat task, rendered in images. A robot agent observes a sequence of actions performed by the human (step 1-3), and receives a quest (step 4). The robot needs to interpret the natural language quest based on physical and social context, and select the relevant object from the environment: the knife on the table in this case. \revise{Currently, the HandMeThat benchmark is released as a text-only environment.}}
  \vspace{-0.5em}
  \label{fig:teaser}
\end{figure}

To collaborate with human partners successfully in complex environments, robots should be able to interpret and follow natural language instructions in contexts. Consider the example shown in \fig{fig:teaser}, a human is preparing fruits for bottling. In the middle of her actions, the human asks a robot agent for help: ``can you hand me that one on the table please?'' The robot needs to correctly interpret the sentence in the current context and interact with objects to accomplish this task.

\revise{Here, the human utterance essentially specifies a subgoal for the robot (getting the knife), derived from her own goal (bottling fruits).  In reality, such subgoal can be under-specified in human utterances, typically for two reasons. First, the human assumes that the robot has knowledge about her goal~\citep{dennett1987intentional,gergely1995taking}. Second, human makes trade-offs between accuracy and efficiency of communication~\cite{grice1975logic,sperber1986relevance,clark1996using}.} While previous benchmarks in similar domains have been primarily focusing on the language grounding of object properties (\eg, ``table''), relations (\eg, ``on''), and planning (\eg, object search and manipulation)~\citep{ALFRED20,srivastava2022behavior}, in this paper, we highlights \revise{the additional challenge for understanding human instructions with {\it ambiguities} (\ie, recognizing the subgoal) based on physical states and human actions and goals.} In this example, the human has just taken fruits out of the refrigerator and is trying to slice them on the countertop. Thus, the object to be retrieved should be the knife.

\revise{Developing a benchmark for resolving ambiguous instructions based on both physical and social environments is challenging in both data collection and automatic evaluation: it is hard to collect the everyday dialogues with corresponding physical states, and it is hard to build automatic evaluation protocols that involve human judgments about robot success.}
In this paper, we present a new benchmark, \benchmark, aiming for a holistic evaluation of language instruction understanding and following in physical and social %
\revise{environments}. 
\revise{We set up the environment using an internal symbolic representation-based simulator, so that we can generate human trajectories and instructions following automated pipelines and build automated evaluation protocols. } 

Each episode in \benchmark contains two stages. In the first stage, the %
\revise{robot} will be watching a human-like agent taking actions towards her internal goal (which is unrevealed to the %
\revise{robot}). At the end of the first stage, we assume that the human needs help from the %
\revise{robot}. Therefore, the %
\revise{robot} receives a (possibly ambiguous) language instruction, which is essentially a subgoal for the %
\revise{human}. In the second stage, the %
\revise{robot} takes actions in the environment to accomplish this subgoal. We evaluate the %
\revise{robot's} performance by his action costs during the second stage and whether %
\revise{his} actions accomplish human's subgoal.

\benchmark contains a diverse set of %
\revise{physical (the objects in the scene) and social (the internal goal of human) information}. Specifically, each \benchmark scene contains 14 locations and typically more than 200 movable objects, which induces a large set of possible actions. The human's internal goal is instantiated from a distribution derived from \behavior~\citep{srivastava2022behavior}. This brings us two advantages. First, since BEHAVIOR tasks are manually annotated, they follow the natural distribution of human household tasks. Second, due to the compositional nature of \behavior, the space of possible goal specifications is enormous. In particular, using the templates we extracted from \behavior, we can instantiate more than 300k distinct tasks. Such diversity introduces important challenges for goal recognition, language understanding, and embodied interaction.

\revise{The key challenge in \benchmark tasks is the recognition of human's subgoal from her historical actions and ambiguous instructions. This resembles three important challenges: recognition of human goals, pragmatic reasoning of natural language, and planning. However, \benchmark is not a simple {\it ensemble} of these three challenges. In goal recognition, the robot needs to consider both human's historical actions as well as the subgoal specified in human utterance. Similarly, the context for pragmatic reasoning consists of both the physical environment and human's internal goals. Furthermore, since our environment is partially observable, the robot can gather additional information through exploration in order to help with goal recognition and pragmatic reasoning. \benchmark integrates these naturally-occurring challenges and serves as a holistic benchmark.}

In this paper, we implement a textual interface to render physical scenes and allow the \revise{robot} agent to use natural language commands to interact with the human and objects. \revise{Rendering in the textual interface bypass the difficulties in visual perception and recognition, allowing us to focus on pragmatic instruction reasoning problem.}
We formulate the learning problem using reinforcement learning, where the %
\revise{robot} receives textual inputs describing the scene and the human actions, and generates natural language commands to accomplish the subgoal specified by the human. For baselines, we compare two groups of baselines. The first set contains a random agent and a heuristics-based agent. The second set contains neural network baselines that are trained with offline and online reinforcement learning algorithms. All learning-based baselines show less than 20\% success rate on our held-out test episodes, suggesting significant room for improvements.

\section{Related Work}
\vspace{-0.5em}
\tbl{tab:comparison} summarizes the comparison between \benchmark and other text-based and vision-based benchmarks in similar domains. \revise{Although \benchmark is only rendered in texts currently, simplifying the perception and recognition problem, we believe that many other aspects we emphasize are critical challenges and it is worth comparing \benchmark with many visual benchmarks.}

\begin{table}[tp!]
\centering\scriptsize
\vspace{-0.5em}
\setlength{\tabcolsep}{5pt}
\begin{tabular}{lcccccccc}
\toprule
       & \begin{tabular}[c]{@{}c@{}}Object\\ Interaction\end{tabular}  
       & \begin{tabular}[c]{@{}c@{}}Goal Space\\ (\#Templates)\end{tabular}
       & \begin{tabular}[c]{@{}c@{}}Common-Sense\\ Goal Prior\end{tabular}  
       & \begin{tabular}[c]{@{}c@{}}Instruction\\ Interpretation\end{tabular}
       & \begin{tabular}[c]{@{}c@{}}Social\\ Reasoning\end{tabular}
       & \begin{tabular}[c]{@{}c@{}}Pragmatic\\ Inference\end{tabular}
       & \begin{tabular}[c]{@{}c@{}}Collaborative\\ Task Completion\end{tabular} \\\midrule
BEHAVIOR \cite{srivastava2022behavior} 
& \Checkmark  
&  100 (100) 
& \Checkmark
& \XSolidBrush 
& \XSolidBrush 
& \XSolidBrush 
& \XSolidBrush \\
ALFRED \cite{ALFRED20} 
& \Checkmark  
& 7,000 (7)
& \XSolidBrush 
& \Checkmark  
& \XSolidBrush
& \XSolidBrush 
& \XSolidBrush \\
Watch-and-Help \cite{puig2021watchandhelp} 
& \Checkmark  
& 5 (5)  
& \XSolidBrush  
& \XSolidBrush 
& \Checkmark  
& \XSolidBrush
& \Checkmark \\
CerealBar \cite{suhr-etal-2019-executing} 
& \Checkmark  
& 1202*
& \XSolidBrush 
& \Checkmark  
& \Checkmark
& \XSolidBrush
& \Checkmark \\
ALFworld \cite{ALFWorld20} 
& \Checkmark 
& 7,000 (7)  
& \XSolidBrush 
& \Checkmark  
& \XSolidBrush
& \XSolidBrush
& \XSolidBrush \\ 
DialFRED \cite{Gao2022DialFREDDA} 
& \Checkmark 
& 36,912 (25)*  
& \XSolidBrush 
& \Checkmark  
& \XSolidBrush
& \XSolidBrush
& \Checkmark \\ 
TEACh \cite{Padmakumar2022TEAChTE}
& \Checkmark 
& 7,000 (7)  
& \XSolidBrush 
& \Checkmark
& \Checkmark
& \XSolidBrush
& \Checkmark \\
Two Body \cite{Jain2019TwoBP}
& \Checkmark 
& 1 (1)  
& \XSolidBrush 
& \XSolidBrush
& \XSolidBrush
& \XSolidBrush
& \Checkmark \\
\midrule
TextWorld \cite{cote2018textworld}
& \Checkmark  
& ---
& \XSolidBrush 
& \Checkmark  
& \XSolidBrush
& \XSolidBrush 
& \XSolidBrush \\
LIGHT \citep{urbanek2019light}
& \XSolidBrush  
& 10,777*
& \XSolidBrush 
& \Checkmark  
& \Checkmark
& \XSolidBrush 
& \XSolidBrush \\
SCONE \citep{fried-etal-2018-unified,long-etal-2016-simpler}
& \XSolidBrush  
& 13,951*
& \XSolidBrush 
& \Checkmark  
& \XSolidBrush 
& \Checkmark 
& \XSolidBrush \\ \midrule
HandMeThat (ours) 
& \Checkmark 
& >300k (69)  
& \Checkmark
& \Checkmark 
& \Checkmark
& \Checkmark  
& \Checkmark  \\ \bottomrule
\end{tabular}
\vspace{0.2cm}
\caption{Comparison between \benchmark and other related benchmarks. *: indicates the number of instructions or dialogues in that dataset, in contrast to explicit goals/tasks. The numbers in the parenthesis in the third column represents the number of goal templates.}
\label{tab:comparison}
\vspace{-2em}
\end{table}

 \xhdr{Household manipulation tasks.}
Robotic manipulation in household environments is an important challenge because it calls for combined research of navigation, object manipulation, and language use. Thus, many household environment simulators and platforms~\citep{puig2018virtualhome, gan2021threedworld, habitat19iccv, szot2021habitat} have been built. As a representative, \behavior~\cite{srivastava2022behavior} \revise{(built on iGibson 2.0~\cite{li2021igibson})} is a physics-based simulator and the only one that contains human-annotated tasks, which reflects a real-world distribution of household activities.
In this paper, we leverage the state representations and task distributions collected by \behavior to study human-robot communication. The task distribution can be viewed as a commonsense prior of human goals and intentions: \eg, human uses knife to cut fruit (in contrast to hammers) and %
\revise{jars} to store them (in contrast to trash bins). By contrast, the original \behavior does not involve any language communication.
The closest benchmark to our work is ALFRED~\citep{ALFRED20} and its successor ALFWorld~\citep{ALFWorld20} \revise{(built on AI2-THOR~\cite{Kolve2017AI2THORAn})}. Both benchmarks use natural language instructions to set up the tasks. However, their task distribution is not diverse: ALFRED has 7,000 different goals in compositional formulas, instantiated from only 7 templates, and they do not follow any real-world task distribution. Furthermore, both ALFRED and ALFWorld do not consider pragmatic reasoning and %
\revise{human actions and goals}.

\xhdr{Goal recognition and social reasoning.}
\revise{Our task formulation is closely related to the literature on goal recognition: inferring the goal of other agents based on their historical actions~\citep{10.5555/1643031.1643119,baker2007goal,Levesque2011,RamirezGeffner09,RamirezGeffner10, series/synthesis/2013Geffner,zhi2020online,Meneguzzi2021ASO}. The most prevalent assumption is the principle of rationality: agents should make (approximately) optimal decisions to achieve their goals, given their beliefs~\citep{dennett1987intentional, gergely1995taking}. Similar to existing work on goal recognition from unstructured data~\citep{8489653}, in \benchmark, the robot does not assume access to domain knowledge. Furthermore, our benchmark \benchmark makes an important extension to the standard setups of goal recognition: besides human actions, \benchmark considers (possibly ambiguous) human instructions that set subgoals for the robot. The objective of the robot is not to fully recover the human's goal, but the subgoal set by the human.}

\revise{Understanding human intentions in embodied environments has been studied in other works.} Watch-and-Help~\citep{puig2021watchandhelp} introduces a non-language goal inference task based on the VirtualHome environment~\cite{puig2018virtualhome}. One critical drawback of their environment is that the goal space is relatively small, and no language interpretation is involved. CerealBar~\cite{suhr-etal-2019-executing} is another dataset for robotic instruction following in a collaborative environment. However, their environment does not reflect real-world priors of goals. \revise{LIGHT~\cite{urbanek2019light} sets up a textual platform for understanding actions and emotes within natural language dialogues, which involves the interpretation of human's internal ideas by focusing on background knowledge such as backstory and personality. Our benchmark, on the contrary, aims at reasoning the internal goal within the planning domain.}

\xhdr{Pragmatic inference.} 
Our work is closely related to the extensive literature on pragmatic inference---the study of context contributes to meaning. This idea is widely applied to referring expression generation tasks~\citep{10.1162/COLI_a_00088, Yu2017AJS} in visual scenes. Our benchmark, similarly, considers human utterance generation for subgoals and considers both physical and social contexts. Related to \benchmark, there is work on integrating pragmatic reasoning and instruction following tasks. Fried~\etal~\cite{fried-etal-2018-unified} proposed a unified pragmatic model for generating and following instructions, which applies pragmatic inference on the navigation dataset SAIL~\cite{MacMahon2006WalkTT} and the semantic parsing dataset SCONE~\cite{long-etal-2016-simpler}. Both tasks focus on modeling the textual contexts (\eg, the previous instructions in a dialog). By contrast, in this paper, we focus on understanding human actions in physical and social \revise{environments}.

\xhdr{\revise{Collaborative Communication.}} \revise{In Two Body Problem~\cite{Jain2019TwoBP}, two agents can communicate in both explicit (through message) and implicit (through perception) ways, in order to efficiently finish a single given task. DialFRED~\cite{Gao2022DialFREDDA}, an extension to ALFRED allows agent to actively ask questions to humans for helpful information. However, the language in DialFRED has no ambiguity, and the questioning is only for information seeking procedure instead of ambiguity resolution.
TEACh~\cite{Padmakumar2022TEAChTE} and CerealBar~\citep{suhr-etal-2019-executing} introduce collaborative tasks where a \textit{Commander} receives a given task and a \textit{Follower} interacts with the environment. In their work, the objective of the {\it Commander} is to accurately describe the task in language to the {\it Follower}. By contrast, our benchmark focuses explicitly on the trade-off between informativeness and communication cost.}

\xhdr{Text-based reinforcement learning.}
We build a textual interface for \benchmark based on gym environment~\cite{brockman2016openai}, which have been used by adventure game environments such as Textworld~\cite{cote2018textworld} and Jericho~\cite{hausknecht19}. In contrast to our benchmark \benchmark, these environments do not involve goal-conditioned learning, and there is no human-robot communication and interactions.

\section{The \benchmark Benchmark}
\vspace{-0.5em}

\begin{figure}[tp]
  \centering\small
  \includegraphics[width=\textwidth]{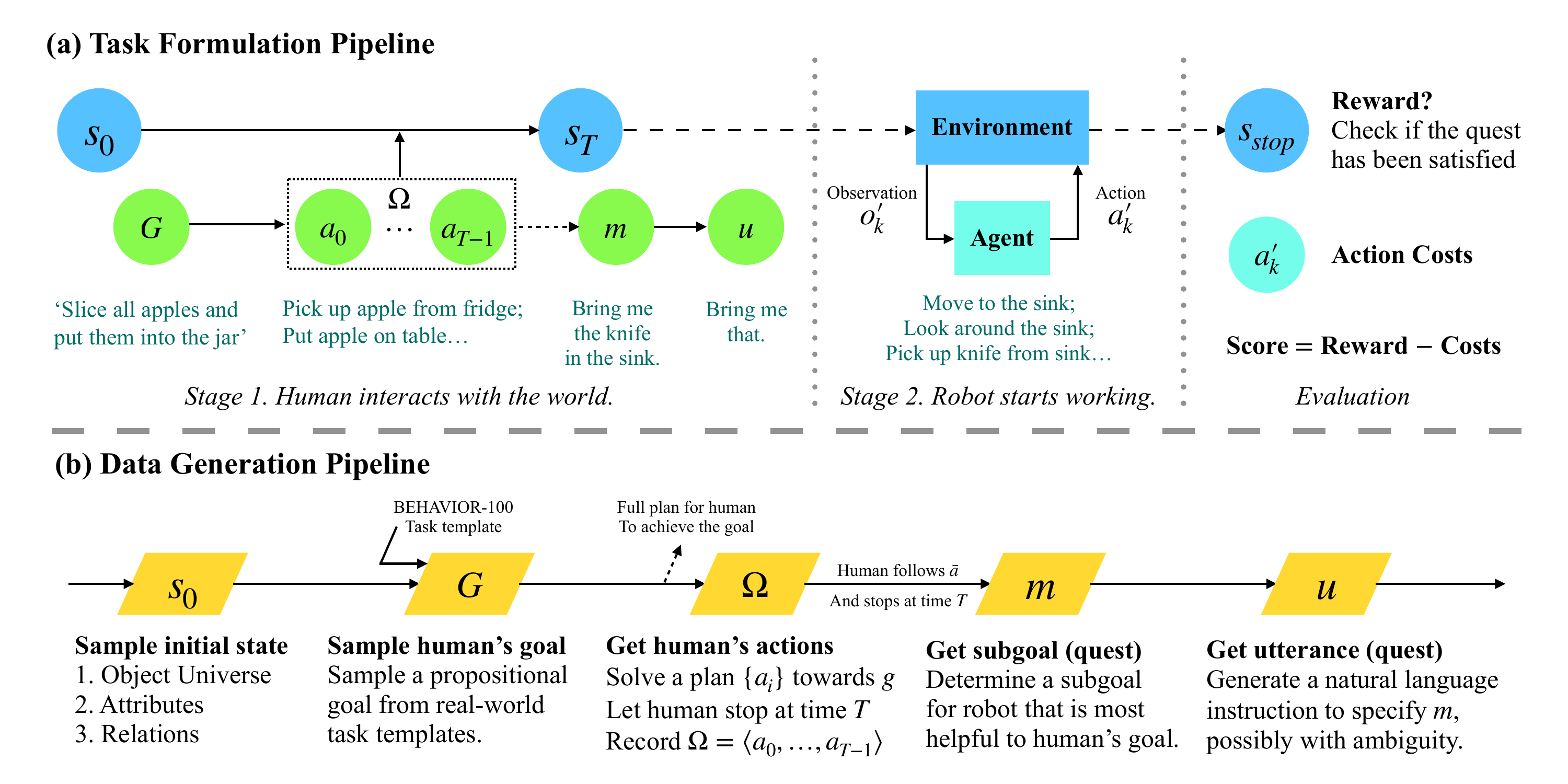}
  \caption{(a) A pipeline for \benchmark task formulation. Stage 1: Human takes $T$ steps from initial state $s_0$ towards a goal $G$, giving a trajectory $\Omega$. At state $s_T$, she generates a subgoal $m$ for robot and utters it as $u$. Stage 2: Robot perceives and acts in the world, following the human's instruction. Evaluation: When the robot stops, we check if human's quest has been satisfied, and count robot's action costs to give a final score.
  (b) A pipeline for \benchmark data generation. 
  We first sample initial state $s_0$, human’s goal $G$, and solve a plan for human to execute. At a randomly sampled step $T$, the human stops and generates a subgoal, including both the internal $m$ and utterance $u$.}
  \label{fig:pipeline}
\end{figure}

\revise{The data generation and evaluation of \benchmark are based on a symbolic representation-based simulator. In this section, we will borrow notations and concepts from classical planning domains~\citep{journals/corr/abs-1106-4561,Helmert2009-HELCFR} to formalize our environment. A domain $\Xi$ is composed of a state space $\gS$, an action space $\gA$, and a transition function $\gamma$.} Each state $s\in \gS$ is represented as an object-centric representation $s = \langle \gU, \gV \rangle$. $\gU$ is the universe of objects in the scene. \revise{$\gV$ is a finite set of state variables. All state variables $v\in \gV$ considered in this paper are binary-valued. They are either unary variables that describe the attributes of each object (\eg, {\tt sliced(apple\#0)}, {\tt dusty(box\#1)}), or binary variables representing the spatial relations between objects (\eg, {\tt in(apple\#1, box\#2)}).} For convenience, we will use $p,r$ to denote the set of attributes and relations separately. In this paper, we only consider two spatial relationships: {\it on} and {\it in}. All object categories and states are inherited from \behavior.

In \benchmark, the action space $\gA=\gA_h\cup \gA_r$ is composed of human actions $\gA_h$ and robot actions $\gA_r$. %
Each action $a\in \gA$ can be represented as $a = \langle \hat{a}, O_\textit{arg} \rangle$, where $\hat{a}$ is an action schema~\cite{fikes1971strips} and $O_\textit{arg}$ is a tuple of object arguments $o_1, \ldots, o_k\in \gU$. For example: {\tt robot-open(cabinet)} means ``robot opens the cabinet,'' and {\tt robot-slice-with-on(human, apple, knife, table)} means ``the robot slices an apple with knife on the table.'' \revise{In STRIPS planning literature, each action $a$ can be considered as a grounded operator, characterized by preconditions and effects, which are logical formula defined over state variables and assignments to state variables, respectively. We leave the detailed definitions of these operators to the supplementary material.} As an example, the action {\tt robot-open(cabinet)} changes the {\tt is-open} property of the cabinet.

\revise{The set of available actions (grounded operators) immediately induces a deterministic and discrete transition function: $\gamma: \gS \times \gA \rightarrow \gS$. $\gamma(s, a)$ computes the outcome state after taking action $a\in \gA$ at state $s \in \gS$}. Meanwhile, we define the cost function $\gC: \gS \times \gA \rightarrow \R$, which computes the effort $\gC(s, a)$ in performing $a$ at state $s$. For simplicity, throughout the paper we assume actions have unit costs. The details of our state representation, action schemas, and action costs can be found in the supplementary material.

\revise{Based on the basic definitions of states, actions, and transition functions, we formally define each \benchmark episode as a tuple $\langle s_0, G, \Omega, s_T, m, u \rangle$. As shown in \fig{fig:pipeline}a, each episode consists of two stage. In the first stage, the human agent takes $T$ steps from the initial state $s_0$ towards goal $G$. The trajectory $\Omega$ is the sequence of human actions, and $s_T$ is the state reached by executing $\Omega$. Then, the human determines a subgoal $m$ in her mind and specifies it through the utterance $u$. In the second stage, the robot observes $\langle\Omega, s_T, u\rangle$, and interacts with the environment.} The performance of the robot will be evaluated by whether his actions accomplish the subgoal $m$ and the total costs of his actions.

\revise{\fig{fig:pipeline}b presents the generation pipeline of \benchmark episodes. We start from randomly sampling the initial world state $s_0$ and the goal $G$ for humans. Second, we generate the human trajectory $\Omega$ assuming that human takes an optimal plan towards the goal (\sect{sec:goal-and-traj}). Next, we generate the subgoal $m$ and utterance $u$ (\sect{sec:meaning-and-utterance}). The key assumption is that the subgoal $m$ is an {\it useful} subgoal towards $G$, while $u$ is generated with a Rational Speech Acts (RSA) model.}

\vspace{-0.5em}
\subsection{Intial States, Goals, and Human Trajectories}
\label{sec:goal-and-traj}
\vspace{-0.5em}

\begin{figure}[tp]
  \centering
  \includegraphics[width=\textwidth]{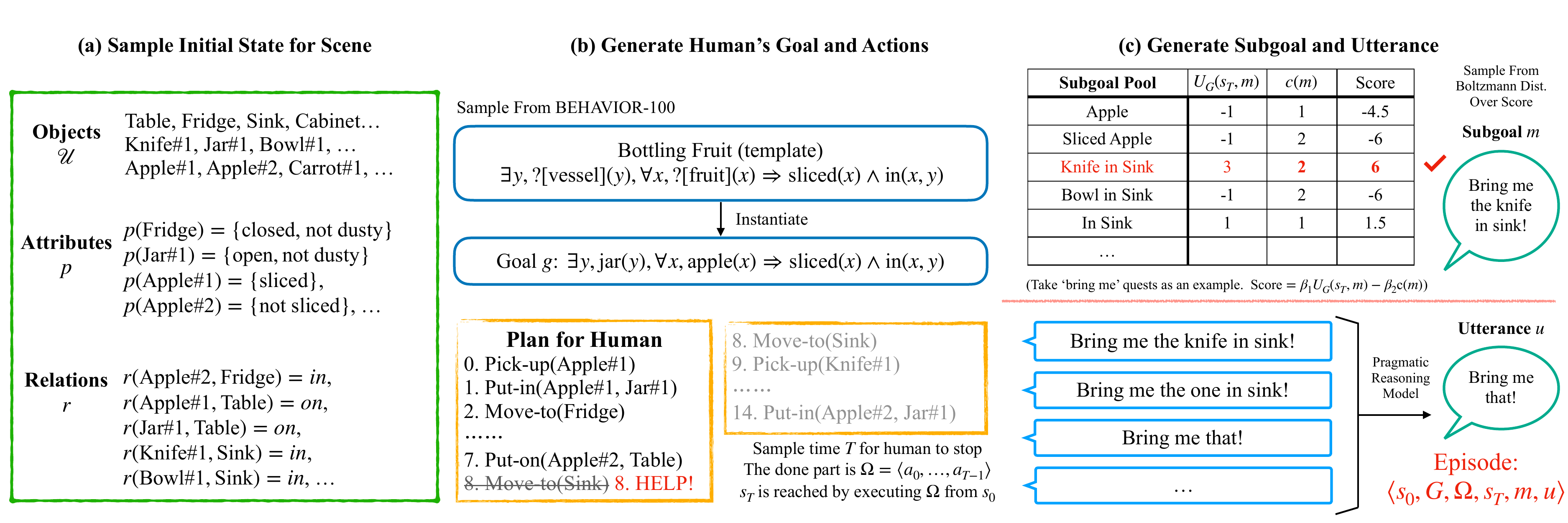}
  \caption{Example of our generated scene, goal, and quest. (a) The sampled initial state. (b) The sampled internal goal from BEHAVIOR-100 templates and the plan. (c) The generated quest in mind based on human's utility and a corresponding utterance generated using the Rational Speech Acts (RSA) model~\citep{frank2012predicting}.}
  \label{fig:detail}
\end{figure}

The initial state \revise{$s_0$} in each \benchmark episode is randomly generated by sampling the number of objects of each object category and then their attributes and spatial relationships. As a result, each scene contains more than 200 entities with diverse attributes, which resembles a typical real household environment. We show one example of generated scenes in \fig{fig:detail}a.

Based on the object-centric state representation, a goal $G$ can be defined as a \revise{first-order-logic} formula over objects in $\gU$. For example, {\tt is-open(cabinet)$\land$on(apple\#0, table)}.
\revise{We say a state $s$ satisfies $G$ if the formula $G$ evaluates to true at $s$.} In each episode, the human generates an internal goal $G$, which is unrevealed to the agent. The goal space $\gG$ is derived from human-annotated household tasks in \behavior, represented using templates of first-order logic statements. For example, the ``bottling fruit'' task (storing sliced fruits) can be formalized as the template shown in \fig{fig:detail}b, and can be instantiated by replacing blanks with concrete object properties.

\revise{Given the domain $\Xi$, the sampled goal $G$, and the initial state $s_0$, we consider a planning task $\Pi=\langle \Xi, s_0, G\rangle$. A solution to a planning task is a sequence of actions $\pi$ that reaches a goal state $G$ starting from the initial state $s_0$ by following the transitions defined in $\Xi$.} We first use a planner to generate a trajectory $\pi = \langle a_0,\ldots,a_n\rangle$ that accomplishes the goal.
Next, to simulate the scenario where the human asks for help, we randomly truncate the trajectory into $T$ steps. \revise{The robot agent observes human actions $\Omega=\langle a_0,\ldots,a_{T-1}\rangle$ and the final state $s_T$.}

\vspace{-0.5em}
\subsection{Subgoals and Utterances}
\label{sec:meaning-and-utterance}
\vspace{-0.5em}

In this paper, we consider three types of subgoals: ``bring-me'' (the robot should deliver an object to the human), ``move-to'' (the robot should move an object to a designated location), and ``change-state'' (the robot should change the state of an object, such as cleaning). Instead of requesting a specific object, the object and target location referred in $m$ is specified using object attributes and relations. That is, instead of specifying ``Bring me the plate with ID 93,'' the subgoal might be ``Bring me a large ceramic one,'' or ``Bring me the plate on the dining table.'' This resembles the real-world use of natural language. 

Formally, we call $m$ a {\it lifted} subgoal. Before explaining that, we define a {\it grounded} subgoal $\textit{mg}$ to be \revise{a propositional logic formula over state variables $\gV$}, \eg, {\tt human-holding(plate\#93)}. Intuitively, a lifted subgoal replaces each concrete object with a dictionary specifier $d$. For example, {\tt plate\#93} is replaced by a specification dictionary {\tt \{size:large, material:ceramic\}}. \revise{The specification dictionary can be translated into a formal first-order-logic formula over objects in $\gU$, as $\exists x.\texttt{ human-holding}(x)\land \texttt{size-large}(x)\land \texttt{material-ceramic}(x)$}

\revise{For each lifted subgoal $m$, we use $A(m)$ to denote the set of all possible grounded subgoals that satisfy $m$. Formally, $\revise{A(m)=\left\{\textit{mg} | \forall s\in \gS. \textit{mg}(s) \rightarrow m(s)\right\}}$, where $\textit{mg}(s),m(s)$ denote the results of evaluating $\textit{mg}, m$ on state $s$, respectively.}

\xhdr{Subgoal generation.} The subgoal generation process is based on the assumption that the human chooses a lifted subgoal that maximizes her internal reward. Such internal reward is composed of two parts: the utility of the subgoal and \revise{the complexity of the subgoal.}

To quantify the utility of the subgoal, we first define a set of helpful functions. Based on $\gamma$ and $\gC$, we denote the optimal (goal-conditional) policy $\pi^*_G(s): \gS \rightarrow \gA$ which computes the first action starting from $s$ to achieve the goal $G$. Furthermore, we define cost-to-go $V^*_G(s)$ as the total cost following $\pi^*_G(s)$ before achieving $G$.

Next, we define a grounded-subgoal transition function $\gamma_{\text{subgoal}}(s, \textit{mg})$. Specifically, let $s_0$ be the current state, $a_0, s_1, a_1, \cdots, a_{k-1}, s_k$ be the unrolling of the optimal policy $\pi^*_{\textit{mg}}(\cdot)$: $a_i = \pi^*_{\textit{mg}}(s_i)$, $s_{i+1} = \gamma(s_{i}, a_{i})$. We define $\gamma_{\text{subgoal}}(s_0, \textit{mg}) = s_k$. Intuitively, this function computes the landing state following the optimal policy $\pi^*$ towards a grounded subgoal $\textit{mg}$. Thus, we can define the utility of a subgoal as:
\begin{align*}
   U_{G}(s_T, m) = V^*_G(s_T) - \frac{1}{|A(m)|}\sum_{\textit{mg} \in A(m)}\left[ V^*_G\left( \gamma_{\text{subgoal}}(s_T, \textit{mg}) \right) \right] 
\end{align*}
where $s_T$ is the state that the human generates the quest. This function quantifies the expected difference between the current cost-to-go and the cost-to-go after the agent accomplishes the subgoal $\textit{mg}$, assuming $\textit{mg}$ is uniformly sampled from $A(m)$. \revise{When $V^*_G\left( \gamma_{\text{subgoal}}(s_T, \textit{mg})\right)<V^*_G(s_T)$, we say that $\textit{mg}$ is \textit{useful} for $G$. We further define $A(G)$ to be the set of all \textit{useful} grounded subgoals for $G$.}

We define the complexity of a subgoal $\textit{c}(m)$ as the number of specifiers in $m$. For example, the subgoal ``bring-me(in sink)'' has only one specifier, and thus its cost is 1. The human chooses a subgoal $m$ based on the following distribution:
\[ P(m | s_T, G) \propto \exp[\beta_1 U_G(s_T, m) - \beta_2 \textit{c}(m)], \]
where $\beta_1 = 3, \beta_2 = 1.5$ are inverse temperature constants of the Boltzmann distribution.

\xhdr{Utterance generation.} \revise{Our utterance generation process follows the Rational Speech Acts (RSA) models. It generates utterance $u$ given the underlying meaning (\ie, the subgoal) $m$.} The utterance $u$ has the same format as $m$, but the object specification dictionaries are less constrained. For example, instead of specifying both the size and the material, the utterance may only contain material specifications {\tt \{material:ceramic\}} (\ie, ``Bring me a ceramic one.'') In the extreme case, the object specifier can be empty, corresponding to the natural language ``Bring me that.'' For convenience, we say $m \subseteq u$ if and only if $A(m) \subseteq A(u)$.

The intuition behind the generation of $u$ follows the RSA model, where the speaker (the human) considers a rational listener (the robot). Formally, the RSA model iteratively defines a sequence of distributions:
\begin{align*}
    & P_{L_0}(m|u)\propto l(u, m)P(m);\\
    P_{S_k}(u|m) \propto \exp[\alpha U_{S_k}(u, m)];\quad&
    P_{L_k}(m|u) \propto P_{S_k}(u|m)P(m).
\end{align*}
Here, for $k \ge 1$, $U_{S_k}$ is a utility function computed as $U_{S_k}(u, m)=\log P_{L_{k-1}}(m|u) - \revise{\alpha'}c(u)$. \revise{$\alpha = 2, \alpha' = 1$ are temperature constants and cost weights.} The literal meaning of a utterance $l(u, m)$ is defined as $\mathbf{1}[A(m)\subseteq A(u)]$, where $\mathbf{1}$ is the indicator function. Intuitively, if $m$ is a finer-grained specification than $u$, $m$ satisfies $u$.
$P(m)$ is the prior distribution of possible meaning $m$, which is defined as $P(m) = P(m | s_T, G)$---essentially assuming that the listener \revise{has correctly recognized} the internal $G$ of the human. We use the same cost function $c(u)$ as $c(m)$, which counts the number of specifiers. In \benchmark, we use a non-trivial $k = 10$ and uses the distribution $P_{S_k}(u|m)$ to sample an utterance $u$ given the meaning $m$. The sampled utterance $u$ is translated into natural language following templates detailed in our supplementary material. \revise{The choice of hyperparameters $\beta_1,\beta_2,\alpha,\alpha',k$ is also discussed in supplementary materials.}

\fig{fig:detail}c shows a concrete example for utterance generation. When the human asks for help at time $T=8$, the remaining part towards her goal is to get a knife and slice the apple. Considering the pool of all possible subgoals, the ones that are related to ``delivering the knife'' have a high utility. Based on the sampled \revise{subgoal} $m$ ``Bring me the knife in the sink,'' we list all possible utterances $u$ that satisfy $m$, by removing a portion of the specifiers (``in-sink'' and ``is-knife'' in this example) and use the RSA model to generate the utterance.

\subsection{Hardness Levels}
\label{sec:hardness-levels}

The \benchmark benchmark holistically evaluates language grounding, goal inference, and pragmatic reasoning. To systematically disentangle these challenges, we split \benchmark into four hardness levels, and the gaps between levels correspond to different challenges.
\revise{Recall that $A(G)$ denotes the set of all \textit{useful} grounded subgoals for goal $G$.} $A(m)$ and $A(u)$ are the sets of all possible grounded subgoals for the lifted subgoal $m$ and the utterance $u$.
Finally, we consider the subgoal derived from pragmatic reasoning. Specifically, denote $r = \arg\max L_{k}(m | u)$ (\ie, the most probable subgoals following RSA given $u$). Denote its corresponding grounding set as $A(r)$. Directly from the definitions, we have $A(m)\subseteq A(G), A(m)\subseteq A(u), A(r)\subseteq A(u)$. The four hardness levels are:

\xhdr{Level 1}: $A(m) = A(u)$. The utterance has no ambiguity. In this case, the instruction understanding task is a pure grounding task: the agent only needs to select the object that satisfies the specification.

\xhdr{Level 2}: $A(m) = A(u)\cap A(G)$. The second level requires social reasoning: the robot can successfully accomplish the task if it can both ground $u$ and infer the human goal $G$ from observations. 

\xhdr{Level 3}: $A(m) = A(r)$. The third level requires all reasoning capabilities combined: the agent need to infer the human goal $G$. Next, it should make pragmatic reasoning based on $G$ and $u$ to derive $r$. 

\xhdr{Level 4}: $A(m) \subset A(r)$. In this case, the human utterance $u$ is inherently ambiguous and can not be resolved even with all reasoning capabilities. In this case, further information gathering is needed. 
\subsection{Text-Based Interactive Interface} 
\label{sec:text-interface}

In this paper, we construct a text-based environment for \benchmark. We implement the environment based on the gym environment interface~\citep{brockman2016openai}. We present an running example of our gym environment in supplementary materials. The initial observation contains the trajectory of human and the language instruction $u$. Every step, the agent can execute robot actions, including moving, examining, and manipulating objects (pick-and-place, heat, \etc). After each action, the agent will receive a new observation containing the object state changes. Each action has the cost $\gC(s, a)$ which is computed using the internal state. Meanwhile, after each action, the environment will check if the subgoal set by the human has been accomplished. When the agent succeeds, it will receive a reward of 100. We set the maximum episode length to 40. Thus, the agent will fail after taking 40 actions.

We consider two environmental settings: fully-observable and partially-observable. Concretely, in the fully-observable setting, the initial observation contains the information of all objects in the environment. By contrast, in the partially-observable setting, the robot can only see the objects at his current location. For receptacles, the robot needs to explicitly open the receptacle to investigate the objects inside. Our text-based environment has a vocabulary of size 250. The average token length of the observation is 860 in the fully observable setting and 140 for the partially observable setting.

\subsection{Challenges in \benchmark Tasks}
\label{sec:challenges}

\revise{The \benchmark resembles three important and co-occurring challenges: goal inference, pragmatic reasoning, and planning. In this section, we briefly discuss the interplay between them and the new challenges set up by \benchmark. First, unlike most relevant literature on goal recognition, which focuses on inferring goals from human actions, \benchmark additionally considers the ambiguous instructions from humans. Furthermore, the target of the task is not to fully recover the internal goal $G$ of the human, but the subgoal $m$ set by instructions, by assuming that $m$ is a useful subgoal towards $G$. Another important challenge in \benchmark is that in real-world deployment of robots, the robot needs to {\it learn from experience} human preferences, such as object placements and dominant hands. A promising direction is to integrate learning algorithms with goal recognition algorithms. For example, Sohrabi~\etal~\citep{sohrabi2016plan} discussed integrating external probability distributions into planning domains for goal recognition. Second, in the pragmatic reasoning tasks of \benchmark, the robot needs to consider both the physical states of objects but also the goals and potential subgoals of humans. Thus, \benchmark can serve as a benchmark towards building language understanding models grounded on not only physical states but also human actions and goals. Finally, as a partially-observable environment, the robot can and should take actions to gather additional information to facilitate his goal recognition and pragmatic reasoning. In our environment, this primarily involves searching for relevant objects for goal recognition and pragmatic reasoning.}

\section{Experiments}
\label{sec:experiments}
\vspace{-0.5em}
We evaluate two sets of methods on \benchmark. The first set of models contains a random agent and a heuristic-based agent. The second set is neural network-based agents trained with offline and online reinforcement learning algorithms.

\subsection{Model Details}

\xhdr{Hand-coded models.} The first model (Random) is an agent that randomly selects a valid action at each time step. The second model (Heuristic) is an agent that heuristically repeats the previous human actions. To be more specific, this model has access to all object states and the underlying logic formula of the utterance. Therefore, it is only applicable in the fully-observable setting. Upon receiving the instruction, the agent generates all possible groundings of instruction and then compares them to the observed human trajectory. The key heuristic of this model is that: human tends to quest for objects that are in the same categories as the previously manipulated ones. The Heuristic model guesses the grounding of the objects in the utterance based on this heuristic. 

\xhdr{Neural network models.} We evaluate two neural network baselines. The first model (Seq2Seq) is based on the sequence-to-sequence model~\cite{10.5555/2969033.2969173} for language modeling, trained with behavior cloning~\citep{bain1995framework} on expert demonstrations. \camera{The expert demonstrations are generated by applying greedy best-first search with FF heuristic~\cite{Hoffmann2001hff}.} \revise{We apply the architecture of ALFWorld Seq2Seq baseline, where the hidden representation of observation strings is obtained by using task description as attention. As a reference, we also provide the performance of Seq2Seq model given extra oracle (goal or subgoal of the human agent in first-order logic format).}
The second model (DRRN) is presented in Jericho~\cite{hausknecht19}, which is a choice-based text-game agent based on Deep Reinforcement Relevance Network (DRRN)~\citep{he2015deep}. It learns a Q function for possible state-action pairs. %
\revise{We also discuss} an offline variation of DRRN. Specifically, instead of actively collecting environmental trajectories based on the current policy, we train the Q function network with expert demonstration trajectories.

\begin{table}[tp]
\centering \small
\begin{tabular}{l cccc}
\toprule
\multirow{2}{*}{Model} & \multicolumn{4}{c}{Partially Observable} \\ \cmidrule{2-5} 
                       & Level 1     & Level 2    & Level 3   & Level 4      \\ \midrule
Human       &   (76\%, 5.1)         &   (52\%, 4.8)             &      (20\%, 5.8)          &     (8\%, 5.0) \\
Random      &  -40.0 (0.0\%, N/A)   &    -39.5 (0.4\%, 16.0)    &     -40.0 (0.0\%, N/A)    &   -40.0 (0.0\%, N/A)  \\
\midrule
Seq2Seq     &  -5.4 (25.5\%, 4.2)   &    -25.8 (10.4\%, 4.1)    &   -34.7 (3.9\%, 4.1)      &    -32.8 (5.3\%, 4.1) \\
\ \ +goal     &  -11.4 (21.0\%, 4.1)  &    -28.8 (8.2\%, 4.1)     &   -30.0 (7.4\%, 4.2)      &    -32.5 (5.5\%, 4.0) \\
\ \ +subgoal  &  -10.1 (22.0\%, 4.1)  &    -18.7 (15.7\%, 4.1)    &   -24.5 (11.4\%, 4.1)     &    -21.9 (13.4\%, 4.1)\\ \midrule
DRRN        &  -40.0 (0.0\%, N/A)   &   -40.0 (0.0\%, N/A).     &    -40.0 (0.0\%, N/A)     &   -40.0 (0.0\%, N/A)  \\
\ \ +offline  &  -40.0 (0.0\%, N/A)   &   -40.0 (0.0\%, N/A).     &    -40.0 (0.0\%, N/A)     &   -40.0 (0.0\%, N/A)  \\

\bottomrule
\end{tabular}
\vspace{0.2cm}
\caption{Experiment results in the partially observable setting. Each model is evaluated on 4 hardness levels with 3 metrics: 1) the average score, 2) the success rate, and 3) the average number of moves in successful episodes.}
\label{tab:baseline-v1-partially}
\vspace{-1em}
\end{table}

\begin{table}[tp]
\centering \small
\begin{tabular}{l cccc}
\toprule
\multirow{2}{*}{Model} & \multicolumn{4}{c}{Fully Observable} \\ \cmidrule{2-5} 
                       & Level 1     & Level 2    & Level 3   & Level 4      \\ \midrule
Human       &  (92\%, 4.4)          &   (80\%, 4.8)             &      (36\%, 4.3)          &     (16\%, 4.5)        \\
Heuristic  &  88.9 (94.9\%, 4.2)   &    -1.9 (28.1\%, 4.4)     &    -23.7 (12.0\%, 4.3)   &   -15.9 (17.8\%, 4.4) \\
Random      &  -40.0 (0.0\%, N/A)   &    -39.8 (0.1\%, 30.0)    &     -40.0 (0.0\%, N/A)    &   -40.0 (0.0\%, N/A)  \\
\midrule
Seq2Seq     &  -9.5 (22.4\%, 4.0)   &    -30.7 (6.8\%, 4.3)    &   -34.6 (4.0\%, 4.3)      &    -32.5 (5.5\%, 4.1) \\
\ \ +goal     &  -15.7 (17.9\%, 4.3)  &    -25.8 (10.4\%, 4.3)     &   -30.5 (7.0\%, 4.4)      &    -29.0 (8.0\%, 4.0) \\
\ \ +subgoal  &  -4.9 (25.9\%, 4.2)  &    -22.0 (13.2\%, 4.2)    &   -21.0 (14.0\%, 4.0)     &    -19.3 (15.3\%, 4.2)\\ \midrule
DRRN        &  -39.8 (0.1\%, 5.0)   &   -39.8 (0.1\%, 10.0)     &    -39.8 (0.1\%, 9.0)     &   -40.0 (0.0\%, N/A)  \\
\ \ +offline  &  -40.0 (0.0\%, N/A)   &   -40.0 (0.0\%, N/A)     &    -40.0 (0.0\%, N/A)     &   -40.0 (0.0\%, N/A)  \\
\bottomrule
\end{tabular}
\vspace{0.2cm}
\caption{Experiment results in the fully observable setting. Each model is evaluated on 4 hardness levels with 3 metrics: 1) the average score, 2) the success rate, and 3) the average number of moves in successful episodes.}
\label{tab:baseline-v1-fully}
\vspace{-1em}
\end{table}

 \vspace{-0.5em}
\subsection{Results}
\vspace{-0.5em}
\label{sec:results}
\revise{The performance of learning models in partially and fully-observable setting is shown in \tbl{tab:baseline-v1-partially} and \tbl{tab:baseline-v1-fully}. } %
We consider three evaluation metrics: 1) the average score of the model; 2) the success rate that the model achieves the goal within limited steps (40); 3) the average number of moves of successful episodes. Scores are averaged on 1,000 episodes. 

The Heuristic model is the best performing model. Its high success rate on Level 1 is mainly because the utterances in level 1 are unambiguous. Since the Heuristic model assumes groundtruth full information about object states and the underlying symbolic representation of utterances, it can successfully interact with the objects. \revise{It also has access to a planner that resolves plans for%
}
\camera{Pick-and-Place tasks.} \camera{To fairly compare it with the other baselines, we only allow the model to perform in a one-trial manner, \ie, the model could only guess once about the specified subgoal and then follow the plan towards it.}
However, when the level gets harder, its success rate drops, and it is increasingly harder for our simple heuristic to select the correct object.

Among the learning-based models, the Seq2Seq model performs decently well on both fully- and partially-observable settings. Its performance also gets worse when the hardness level increases. Note that the average number of moves when Seq2Seq succeeds is low and stable. It is possibly because the model has learned a fixed template for many pick-and-place tasks that can be accomplished in 4 or 5 steps. There's a large gap between Seq2Seq and the Heuristic model. Apart from the extra information provided to the Heuristic model, there are several other challenges for the Seq2Seq model. First, human goal space is enormous. Thus, a similar scene and instruction may not correspond to the same goal. Directly learning a mapping from observations to actions can be prone to spurious contexts. %
\revise{Interestingly, for level 1 and 2, the Seq2Seq model shows slightly better performance in the partially-observable setting, though the expert demonstration is the same. We attribute this to the network capacity issue. Compared with the partially-observable model, the fully-observable model uses a separate encoder for the full observation strings. Since we keep the model size to the same, the partially-observable model has doubled the parameters for encoding task-relevant information such as the objects at the current location. When the models are trained with extra oracle information (goal or subgoal of human), all the performances except level 1 significantly improve. This highlights that inferring human goals and subgoals is helpful in resolving ambiguous instructions.}

The reinforcement learning-based DRRN model, by contrast, totally fails in both settings, yielding a similar performance as the Random agent. This is primarily due to the large action space for the model. At each step, there are 15 to 30 valid actions, and the agent gets a sparse reward only when it successfully accomplishes the task. Therefore, in practice, we see that the random exploration strategy in the beginning stage of DRRN's training struggles to find positive rewards. 

\revise{As a reference, we present human performance on \benchmark. We collect human performance on each hardness level. We first introduce human subjects to the benchmark settings including possible locations, objects, and tasks. Next, human subjects interact with our textual interface. For both fully- and partially-observable settings, we collect 25 episodes of human performance for each hardness level. Human subjects are only allowed to choose one particular object to manipulate, and stop immediately after execution no matter if the goal is reached. The results show that our human subjects can perform well on tasks in level 1 and 2. In level 3, the inference of speaker intention gets more difficult. As a result, the performance drops significantly, but still outperforms our learning baselines. Tasks in level 4 are intrinsically under-specified. Thus, the human subjects perform worse than level 3. Note that our environment supports robots asking additional clarification questions to the simulated human in order to resolve ambiguities. However, since no current learning models are not capable of actively asking questions, for a fair comparison, we do not enable this option in human experiments.}

\camera{In general, the performance of all baselines is low on our \benchmark benchmark, especially for level 3 and level 4 tasks. There are two main reasons for this observation. First, there exists a large number of candidate objects in the environment. In particular, when there are multiple objects with similar categories (\eg, different kinds of drinks), it can be hard for models to select the objects that are relevant to human's goals and ambiguous instructions. Second, a portion of the level 3 and level 4 tasks are ``intrinsically'' unsolvable. That is, the information in human's historical actions and instructions may not be sufficient to accurately select the relevant objects. This is a desired feature for level 4 tasks, where further information gathering is required (see our supplementary for more discussions on the extension of the dataset). For level 3 tasks, such information insufficiency occurs as an artifact of the Rational Speech-Acts model. Specifically, we are using the groundtruth goal specification when computing listener's prior. Theoretically, we should replace this groundtruth information with a distribution of possible human goals that can be inferred from the historical actions and instructions, for example, by leveraging inverse planning algorithms~\citep{baker2007goal}. However, this requires an intractable computation due to the large size of our goal space. Therefore, we choose to implement the current RSA computation based on groundtruth goal information.}

\section{HandMeThat Dataset Version 2}
\vspace{-1em}
There are two main issues in the original released version of HandMeThat. First, the total amount of data is small, containing only 120 episodes on average for each goal in the training set. Second, some of the instances are intrinsically unsolvable, even for humans. To be more specific, most of the goals contain several subgoals. After a subset of the subgoals has been achieved, human instruction can be insufficient to specify the concrete next subgoal.

Therefore, we have released an updated version of the dataset, trying to address these problems, so that we can better focus on the core challenges stated in Section~\ref{sec:challenges}: goal inference, pragmatic reasoning, and planning. In summary, we make the following changes: restrict the task to a smaller set of goals, generate much more data pieces, and add a new heuristic for human trajectory generation to address specification ambiguities.

\vspace{-0.5em}
\subsection{Revision Details}
\vspace{-0.5em}

\xhdr{Task scope.} 
For the ease of generating a larger amount of trajectories and controlling the ambiguities of the tasks, in version 2, we restrict ourselves to a smaller subset of goals and only ``bring-me''-typed instructions (see Section~\ref{sec:meaning-and-utterance}). This smaller set consists of 25 goal templates that only involve ``pick-and-place'' actions (but not ``change-state'' actions). That is, to accomplish such goals, we only need to move several objects to their target positions without changing their states by cleaning, slicing, or heating them. This set of goals is sufficient to address the difficulty of goal recognition, considering the large sampling space of world states. More details are provided in Appendix~\ref{sec:data-goal-space}.

\xhdr{Data amount.} We increase the size of the dataset by 40 times. Concretely, the version 2 dataset contains 116,146 episodes in total, with approximately 5,000 episodes for each goal template. The numbers of episodes in different hardness levels are 4008, 78488, 28343, and 5307, respectively. As there are much more data for higher hardness levels, we can better evaluate model performances on goal recognition and pragmatic reasoning. We follow the original dataset to split the dataset into training, validation, and test folds, as shown in \tbl{tab:split-v2}. %

\begin{wraptable}{r}{0.5\textwidth}
\centering \small
\vspace{-1em}
\setlength{\tabcolsep}{5pt}
\begin{tabular}{l cccc}
\toprule
& \textbf{Level 1} & \textbf{Level 2} & \textbf{Level 3}  & \textbf{Level 4} \\ \midrule
Train &3620   &70583   &25488   &4804  \\
Validation &207   &3937   &1433   &247  \\
Test   &181   &3968   &1422   &256  \\
 \bottomrule
\end{tabular}
\vspace{0.2cm}
\caption{The distribution of 116,146 instances in \benchmark (Version 2).}
\label{tab:split-v2}
\vspace{-1em}
\end{wraptable}

\xhdr{Data generation.} Recall that in the original version of dataset, there are cases where it is intrinsically ambiguous to predict the next subgoal based on the human trajectories. For example, even if the robot knows the human wants a fruit and a drink for a picnic, it cannot distinguish between milk and juice if the previous trajectory only involves actions on fruits. Therefore, we add new heuristic rules to generate human trajectories (\ie, to decide the truncation positions) so that the subgoal underlying the human instruction can be predicted based on the previous human actions. In addition, our dataset includes a new annotated label ``subgoal'' for each episode, which is the first-order logic statement from the full goal statement that is associated with the current human intention. Such information can be used as additional supervision for training the goal recognition module.

\vspace{-0.5em}
\subsection{Experiment Results}
\vspace{-0.5em}
We replicate the experiments in Section~\ref{sec:experiments} on the new version. The performance of baseline models in both partially and fully-observable settings are shown in \tbl{tab:baseline-v2-partially} and \tbl{tab:baseline-v2-fully}. Similarly, we consider three evaluation
metrics: 1) the average score of the model; 2) the success rate that the model achieves the goal within
limited steps (40); 3) the average number of moves of successful episodes. We briefly compare the new results with the ones on the original version (Section~\ref{sec:results}). (The learning models will use up a limited number of steps when they fail, whereas humans and the Heuristic model will choose to stop early, so the total scores in the experimental results are not directly comparable.)

\begin{table}[tp]
\centering \small
\begin{tabular}{l cccc}
\toprule
\multirow{2}{*}{Model} & \multicolumn{4}{c}{Partially Observable} \\ \cmidrule{2-5} 
                       & Level 1     & Level 2    & Level 3   & Level 4      \\ \midrule
Human       &   (100\%, 7.8)         &   (80\%, 7.5)             &      (40\%, 8.4)          &     (30\%, 8.0) \\
Random      &  -40.0 (0.0\%, N/A)   &    -39.5 (0.4\%, 16.0)    &     -40.0 (0.0\%, N/A)    &   -40.0 (0.0\%, N/A)  \\
\midrule
Seq2Seq     &  1.3 (30.4\%, 4.0)   &    2.4 (31.2\%, 4.0)    &   -17.7 (16.4\%, 4.1)      &    -23.7 (12.0\%, 4.0) \\
\ \ +goal     &  0.6 (29.8\%, 4.0)  &    3.0 (31.6\%, 4.0)     &   -9.6 (22.4\%, 4.1)      &    -9.6 (22.4\%, 4.1) \\
\ \ +subgoal  &  0.6 (29.8\%, 4.0)  &    14.4 (40.0\%, 4.0)    &   -1.9 (28.0\%, 4.1)     &    -1.4 (28.4\%, 4.0)\\ \midrule
DRRN        &  -40.0 (0.0\%, N/A)   &   -40.0 (0.0\%, N/A).     &    -40.0 (0.0\%, N/A)     &   -40.0 (0.0\%, N/A)  \\
\ \ +offline  &  -40.0 (0.0\%, N/A)   &   -40.0 (0.0\%, N/A).     &    -40.0 (0.0\%, N/A)     &   -40.0 (0.0\%, N/A)  \\

\bottomrule
\end{tabular}
\vspace{0.2cm}
\caption{Experiment results for version 2 dataset in the partially observable setting. Each model is evaluated on 4 hardness levels with 3 metrics: 1) the average score, 2) the success rate, and 3) the average number of moves in successful episodes. This table repeats the result of Table~\ref{tab:baseline-v1-partially} on the version 2 dataset.}
\label{tab:baseline-v2-partially}
\vspace{-1em}
\end{table}

\begin{table}[tp]
\centering \small
\begin{tabular}{l cccc}
\toprule
\multirow{2}{*}{Model} & \multicolumn{4}{c}{Fully Observable} \\ \cmidrule{2-5} 
                       & Level 1     & Level 2    & Level 3   & Level 4      \\ \midrule
Human       &   (100\%, 4.1)         &   (80\%, 4.2)             &      (40\%, 5.2)          &     (30\%, 4.5) \\
Heuristic  &  95.8 (100.0\%, 4.2)   &    59.5 (64.0\%, 4.1)     &    34.5 (39.2\%, 4.1)   &   24.4 (29.2\%, 4.1) \\
Random      &  -40.0 (0.0\%, N/A)   &    -39.8 (0.1\%, 30.0)    &     -40.0 (0.0\%, N/A)    &   -40.0 (0.0\%, N/A)  \\
\midrule
Seq2Seq     &  1.3 (30.4\%, 4.0)   &    -0.8 (28.8\%, 4.0)    &   -22.6 (12.8\%, 4.0)      &    -19.9 (14.8\%, 4.0) \\
\ \ +goal     &  -4.7 (26.0\%, 4.0)  &    7.3 (34.8\%, 4.1)     &   -6.9 (24.4\%, 4.1)      &    -6.9 (24.4\%, 4.1) \\
\ \ +subgoal  &  53.8 (69.1\%, 4.2)  &    51.3 (67.2\%, 4.1)    &   1.3 (30.4\%, 4.1)     &    3.0 (31.6\%, 4.1)\\ \midrule
DRRN       &  -40.0 (0.0\%, N/A)   &   -40.0 (0.0\%, N/A)     &    -40.0 (0.0\%, N/A)     &   -40.0 (0.0\%, N/A)  \\
\ \ +offline  &  -40.0 (0.0\%, N/A)   &   -40.0 (0.0\%, N/A)     &    -40.0 (0.0\%, N/A)     &   -40.0 (0.0\%, N/A)  \\
\bottomrule
\end{tabular}
\vspace{0.2cm}
\caption{Experiment results for Version 2 dataset in the fully observable setting. Each model is evaluated on 4 hardness levels with 3 metrics: 1) the average score, 2) the success rate, and 3) the average number of moves in successful episodes. This table repeats the result of Table~\ref{tab:baseline-v1-fully} on the version 2 dataset.}
\label{tab:baseline-v2-fully}
\vspace{-1em}
\end{table}

The Heuristic model is the best-performing model, and its experimental results are comparable to human performance. Among the learning-based models, the accuracy of the Seq2Seq model increases significantly, especially on higher hardness levels. We attribute this to the reduction in ambiguities in the dataset as well as the increased amount of training data. With extra oracle information (goal or subgoal of human), the performance of the Seq2Seq model gets further improvements. The reinforcement learning-based DRRN model, by contrast, still totally fails, yielding a similar performance as the random agent.

By looking at the output of the model, we find that the Seq2Seq model is able to perform well on a small number of goals. Thus, its accuracy is affected by the frequency of occurrence of these goals. When generating this dataset, we did not require an equal probability of occurrence for each goal on {\it each} hardness level. In particular, the episodes in level 2 contain many tasks with relatively ``simple'' goals. 
\section{Conclusion}
\vspace{-0.5em}
We introduce \benchmark, a benchmark for evaluating instruction understanding and following within physical and social contexts. \benchmark requires the ability to resolve possible {\it ambiguities} in human instructions based on both the physical states of scene objects and human's internal long-term goal. We present a textual interface for \benchmark and evaluate several baselines on it. The experiment results suggest that \benchmark introduces important challenges for human-robot interaction models. We hope that \benchmark will motivate the development of more robust systems that can accomplish Human-AI communication in complex physical and social environments. \revise{Our benchmark focuses on resolving ambiguous instruction, but there exist more uncertainty issues in real-world Human-Robot Interaction (HRI) tasks (\eg non-verbal communication~\citep{chen2021yourefit}). Thus, in future work, we may render \benchmark into a visual interface and add back the other challenges within the visual domain.} Another exciting challenge will be to extend the current \benchmark to multi-round human-robot communications, essentially forming a physically and socially grounded dialog. Future research may consider joint pragmatics reasoning of physical, social, and textual contexts, and information gathering through language.

\xhdr{Acknowledgements.} We thank the anonymous reviewers for their constructive comments and suggestions. We thank our friends and colleagues for participating in the human experiments and providing helpful feedback. This work is in part supported by ONR MURI (N00014-13-1-0333), the Center
for Brain, Minds, and Machines (CBMM, funded by NSF STC award CCF-1231216), the MIT Quest for Intelligence, and MIT–IBM AI Lab. Any opinions, findings, and conclusions or recommendations expressed in
this material are those of the authors and do not necessarily reflect the views of our sponsors.

{\small
\bibliographystyle{unsrt}
\bibliography{reference}
}

\newpage

\appendix

\begin{center}
    \bf \Large Supplementary Material for \benchmark: Human-Robot Communication in Physical and Social Environments
\end{center}
\vspace{1em}

The generated episodes, the gym-style environment, and codes for reproducing baseline results are available on our website: \url{http://handmethat.csail.mit.edu/}. This supplementary material document is organized as the following. In \sect{sec:task-design}, we provide the detailed information for \benchmark data generation and its textual interface. In \sect{sec:statistics}, we summarize the statistics of the (version 1) dataset. In \sect{sec:extended-experiments} we present the experimental details including the baseline implementations, the full results with error bars, and the performance of Offline-DRRN, an offline variant of the DRRN model. \sect{sec:human-interaction} discusses the possibility of introducing human-robot interaction to \benchmark. 
\section{Task Design}
\label{sec:task-design}
\subsection{Object Space}
Recall that \benchmark uses an object-centric representation for states. Here, we present all the object categories, attributes, and how the object universe is sampled.

\xhdr{Object hierarchy.} 
\benchmark classifies all categories into 5 classes: location, receptacle, food, tool, and thing. ``Location'' consists of all non-movable entities. They are the positions that an agent can ``move to.'' ``Receptacle'' consists of all movable containers. The latter three classes are ordinary (non-receptacle) movable physical entities.

Each class (except for ``location'') is composed of multiple subclasses, and each subclass contains several object categories. In total, there are 155 object categories. We list the hierarchies for all the five classes in \tbl{tab:hierarchy}. In a scene, there could be multiple objects within a same category, labeled with different IDs.
\begin{table}[tp!]
\centering\small
\vspace{-0.5em}
\setlength{\tabcolsep}{5pt}
\begin{tabular}{lll}
\toprule
        \textbf{Class} & \textbf{Subclass} & \textbf{Category}   \\ \midrule
\multirow{2}{*}{location}  
&(has-ontop)
&\begin{tabular}[l]{@{}l@{}}
floor, countertop, sofa, bed, \\
stove, table, shelf, toilet
\end{tabular}\\ \cmidrule{2-3}

&(has-inside)
&\begin{tabular}[l]{@{}l@{}}
cabinet, bathtub, microwave, oven, \\
dishwasher, refrigerator, sink, pool
\end{tabular}\\\midrule

\multirow{13}{*}{receptacle}
& furniture & highchair, chair, seat\\ \cmidrule{2-3}
& vessel    & bottle, jar, kettle, caldron\\ \cmidrule{2-3}
& tableware & bowl, mug, plate, dish, cup\\ \cmidrule{2-3}
& utensil   & saucepan, pan, casserole\\ \cmidrule{2-3}
& bag       & duffel bag, sack, backpack, briefcase\\ \cmidrule{2-3}
& bucket & bucket\\ 
& tray & tray\\
& basket & basket\\
& box & box\\
& package & package\\
& ashcan & ashcan\\
& Xmas stocking & Xmas stocking\\
& Xmas tree & Xmas tree\\ \midrule

\multirow{10}{*}{food}
& fruit
& \begin{tabular}[l]{@{}l@{}}
apple, banana, melon, grape, lemon, orange, peach, \\
strawberry, raspberry, date, olive, chestnut
\end{tabular}\\
\cmidrule{2-3}
& vegetable
& \begin{tabular}[l]{@{}l@{}}
carrot, radish, tomato, broccoli, \\
mushroom, onion, lettuce, pumpkin
\end{tabular}\\
\cmidrule{2-3}
& drink & pop, beer, juice, water, milk\\
\cmidrule{2-3}
& protein & beef, chicken, pork, fish, egg\\
\cmidrule{2-3}
& flavorer & catsup, sauce, parsley, tea bag, suger, vegetable oil\\
\cmidrule{2-3}
& baked food & cracker, bread, cookie, cake\\
\cmidrule{2-3}
& snack & chip, hamburger, sandwich, candy\\
\cmidrule{2-3}
& prepared food & oatmeal, sushi, salad, soup, pasta\\

\multirow{10}{*}{tool}
& metal tool
& carving knife, hammer, screwdriver, scraper, saw\\
\cmidrule{2-3}
& electric equipment
& printer, scanner, facsimile, modem\\
\cmidrule{2-3}
& electrical device 
& calculator, headset, earphone, mouse, alarm\\
\cmidrule{2-3}
& toiletry & toothbrush, perfume, makeup\\
\cmidrule{2-3}
& writing tool & highlighter, marker, pen, pencil\\
\cmidrule{2-3}
& piece of cloth & dishtower, hand towel, rag\\
\cmidrule{2-3}
& cleaning tool & scrub brush, broom, vacuum\\
\cmidrule{2-3}
& cleansing & soap, shampoo, detergent, toothpaste\\
\cmidrule{2-3}
& cutlery & fork, spoon, knife\\
\cmidrule{2-3}
& illumination tool & lamp, candle\\ \midrule

\multirow{8}{*}{thing}
& decoration
& necklace, bracelet, jewelry, bow, wreath, ribbon\\
\cmidrule{2-3}
& paper product
& \begin{tabular}[l]{@{}l@{}}
hardback, notebook, book, newspaper, \\
painting, pad, document
\end{tabular}\\
\cmidrule{2-3}
& footwear
& gym shoe, sandal, shoe, sock\\
\cmidrule{2-3}
& headwear
& hat, sunglass\\
\cmidrule{2-3}
& clothing & shirt, sweater, underwear, apparel\\
\cmidrule{2-3}
& building materials & tile, plywood\\
\cmidrule{2-3}
& plaything & cube, ball\\

\bottomrule
\end{tabular}
\vspace{0.2cm}
\caption{Object hierarchy designed for \benchmark based on \behavior.}
\label{tab:hierarchy}
\end{table}

\xhdr{Attributes and relations.}
Each object category is also associated with several attributes. There are 2 static attributes and 8 non-static attributes. The static ones are \textit{size} and \textit{color}, which include \{large, small\} and \{red, green, blue\}, respectively. The non-static attributes are all Boolean-valued, including \textit{open, cooked, frozen, dusty, stained, sliced, soaked, toggled}. Each attribute is associated with a meta-predicate: {\it has-X} or {\it X-able} (\eg, {\it has-size} and {\it openable}), indicating whether a object (of a certain category) can be assigned  assigned with that attribute.

For relations, we only consider spatial relations between object pairs, \ie, ``inside'' and ``on top of.'' Each movable object can be in/on a ``location'' object, and each non-receptacle movable object can be in/on a ``receptacle'' object. Note that we only explicitly represent \textit{direct} spatial relations, that is, when \textit{apple\#1 is in plate\#1} and {\it plate\#1 is on table\#1}, we do not explicitly represent \textit{apple\#1 is on table\#1}. This representation brings us the advantage that when the plate is moved to another place, we do not need to manually re-assign the relationship between apple\#1 and table\#1. There are two additional meta-properties, associated with location and receptacle categories: ``has-inside'' and ``has-ontop,'', indicating whether the object of the corresponding category can hold another object inside or on top of it. 
We include all details in \tbl{tab:attributes}.

\begin{table}[tp!]
\centering\small
\vspace{-0.5em}
\setlength{\tabcolsep}{5pt}
\begin{tabular}{ll}
\toprule
        \textbf{Meta-Properties}
       & \textbf{Categories} \\ \midrule
has-inside
& 
\begin{tabular}[c]{@{}l@{}}
cabinet, bathtub, microwave, oven, dishwasher, refrigerator, sink, pool, \\
vessel, tableware, utensil, bag, basket, box, package, ashcan, bucket, Xmas stocking
\end{tabular}  \\ \midrule
has-ontop
& 
\begin{tabular}[c]{@{}l@{}}
floor, countertop, sofa, bed, stove, table, shelf, toilet \\
furniture, tray, Xmas tree
\end{tabular}  \\
\midrule
has-size
& 
tableware tray box package ashcan  \\
has-color
& 
furniture vessel bag basket box package  \\
\midrule
openable
& 
cabinet,  microwave,  oven,  dishwasher, refrigerator, vessel, bag, box, package \\
toggleable
& 
microwave, oven dishwasher, refrigerator, stove, sink, electric equipment\\
cookable
&
food\\
freezable
&
food\\
sliceable
&
fruit, vegetable, protein\\
dustyable
&
location, receptacle, thing\\
stainable
&
location\\
soakable
& piece of cloth, clothing\\

\bottomrule
\end{tabular}
\vspace{0.2cm}
\caption{All meta-properties in \benchmark and their corresponding categories. ``has-inside'' and ``has-ontop'' indicate whether the objects in this category can hold other objects in the corresponding relation; ``has-size'' and ``has-color'' indicate whether the objects in this category will be assigned with size or color; the remaining properties (X-able) indicate whether objects in this category can be manipulated through corresponding verb. The category list contains class/subclass names: all categories derived from the listed class/subclass are associated with the corresponding meta-property.}
\label{tab:attributes}
\end{table}

\xhdr{Valid actions.}
To interact with the world, an agent can perform a variety of actions. We present all the action schemas in \tbl{tab:actions} written in PDDL format~\citep{fikes1971strips}. 
\revise{Formally, each action should be defined as a collection of preconditions and effects, which represents the change in state variables. But here in the table we only introduce the intuitive definitions in Semantics column. Please refer to our code at \url{http://handmethat.csail.mit.edu/} for a complete definition of STRIPS operators in HandMeThat.}
Note that some of the actions are only for robot agents.
Additionally, there are constraints specified for a partial set of actions, as shown in \tbl{tab:restrictions}.

\begin{table}[tp!]
\centering\small
\vspace{-0.5em}
\setlength{\tabcolsep}{1pt}
\begin{tabular}{ll}
\toprule
        \textbf{Action Name and Parameters} & \textbf{Semantics} \\ \midrule
\mycell{{\bf Action: move}\\{\bf Args:} ?a - agent ?from - location ?to - location}
& Agent [?a] moves from [?from] to [?to]. \\
\midrule
\mycell{{\bf Action: pick-up-at-loc}\\{\bf Args:} ?a - agent ?obj - object ?loc - location}
& Agent [?a] picks up [?obj] at [?loc].\\
\mycell{{\bf Action: pick-up-from-rec-at-loc}\\{\bf Args:} ?a - agent ?obj - object ?rec - receptacle ?loc - location}
& Agent [?a] picks up [?obj] from [?rec] at [?loc].\\
\midrule
\mycell{{\bf Action: put-inside-loc}\\{\bf Args:} ?a - agent ?obj - object ?loc - location}
& Agent [?a] puts [?obj] into [?loc].\\
\mycell{{\bf Action: put-ontop-loc}\\{\bf Args:} ?a - agent ?obj - object ?loc - location}
& Agent [?a] puts [?obj] onto [?loc].\\
\mycell{{\bf Action: put-inside-rec-at-loc}\\{\bf Args:} ?a - agent ?obj - object ?rec - receptacle ?loc - location}
& Agent [?a] puts [?obj] into [?rec] at [?loc].\\
\mycell{{\bf Action: put-ontop-loc-at-loc}\\{\bf Args:} ?a - agent ?obj - object ?rec - receptacle ?loc - location}
& Agent [?a] puts [?obj] onto [?rec] at [?loc].\\
\midrule
\mycell{{\bf Action: open-loc}\\{\bf Args:} ?a - agent ?loc - location}
& Agent [?a] opens [?loc].\\
\mycell{{\bf Action: close-loc}\\{\bf Args:} ?a - agent ?loc - location}
& Agent [?a] closes [?loc].\\
\mycell{{\bf Action: open-rec-at-loc}\\{\bf Args:} ?a - agent ?rec - receptacle ?loc - location}
& Agent [?a] opens [?rec] at [?loc].\\
\mycell{{\bf Action: close-rec-at-loc}\\{\bf Args:} ?a - agent ?rec - receptacle ?loc - location}
& Agent [?a] closes [?rec] at [?loc].\\
\midrule
\mycell{{\bf Action: toggle-on-loc}\\{\bf Args:} ?a - agent ?loc - location}
& Agent [?a] toggles on [?loc].\\
\mycell{{\bf Action: toggle-off-loc}\\{\bf Args:} ?a - agent ?loc - location}
& Agent [?a] toggles off [?loc].\\
\mycell{{\bf Action: toggle-on-obj-at-loc}\\{\bf Args:} ?a - agent ?obj - object ?loc - location}
& Agent [?a] toggles on [?obj] at [?loc].\\
\mycell{{\bf Action: toggle-off-obj-at-loc}\\{\bf Args:} ?a - agent ?obj - object ?loc - location}
& Agent [?a] toggles off [?obj] at [?loc].\\
\midrule
\mycell{{\bf Action: heat-obj}\\{\bf Args:} ?a - agent ?obj - object ?loc - location}
& Agent [?a] heats up [?obj] at [?loc].\\
\mycell{{\bf Action: cool-obj}\\{\bf Args:} ?a - agent ?obj - object ?loc - location}
& Agent [?a] cools down [?obj] at [?loc].\\
\mycell{{\bf Action: soak-obj}\\{\bf Args:} ?a - agent ?obj - object ?loc - location}
& Agent [?a] makes [?obj] soaked at [?loc].\\
\midrule
\mycell{{\bf Action: slice-obj}\\{\bf Args:} ?a - agent ?obj - object ?tool - tool ?loc - location}
& Agent [?a] slices up [?obj] with [?tool] at [?loc].\\
\midrule
\mycell{{\bf Action: clean-obj-at-loc}\\{\bf Args:} ?a - agent ?obj - object ?tool - tool ?loc - location}
& Agent [?a] cleans up [?obj] with [?tool] at [?loc].\\
\mycell{{\bf Action: clean-loc}\\{\bf Args:} ?a - agent ?tool - tool ?loc - location}
& Agent [?a] cleans up [?loc] with [?tool].\\
\midrule
\mycell{{\bf Action: bring-to-human}\\{\bf Args:} ?r - robot ?obj - object ?h - human}
& Robot [?r] brings [?obj] to human [?h].\\
\mycell{{\bf Action: take-from-human}\\{\bf Args:} ?r - robot ?obj - object ?h - human}
& Robot [?r] takes [?obj] from human [?h].\\
\bottomrule
\end{tabular}
\vspace{0.2cm}
\caption{All actions in \benchmark. If the action schema has an argument of type ``agent,'' the action can be executed by either the human or the robot.}
\label{tab:actions}
\end{table}

\begin{table}[tp!]
\centering\small
\vspace{-0.5em}
\setlength{\tabcolsep}{5pt}
\begin{tabular}{ll}
\toprule
\textbf{Action Name} & \textbf{Constraints} \\ \midrule
\begin{tabular}[l]{@{}l@{}}
pick-up\\
put-inside \end{tabular}
& The receptacle/location must be open if it's openable. \\
\midrule
heat
& The agent can only heat things at microwave, oven and stove.\\
cool
& The agent can only cool things at refrigerator.\\
soak
& The agent can only soak things at sink.\\
\midrule
slice
& The agent can only slice things with knife.\\
\midrule
clean
& \begin{tabular}[l]{@{}l@{}}
The agent can only clean things with suitable cleaning tool\\
within rag, dishtowel, hand towel, scrub brush, vacuum, broom. \end{tabular}\\
\bottomrule
\end{tabular}
\vspace{0.2cm}
\caption{Action constraints in \benchmark.}
\label{tab:restrictions}
\end{table}

\xhdr{Initial state sampling.}
The object universe is sampled in four steps. First, we create exactly one instance of all categories in the ``location'' category. Second, we sample the number of objects in each category (up to 3). We make sure that there is at least one object in each subclass. The objects that belong to the same category are labeled by different IDs. To this end, we have generated a massive number of objects.

The third step is to sample the positions for all movable objects. We list all the valid positions for categories in each subclass in \tbl{tab:positions}, to reflect the distribution of real-world object placements. The position is uniformly sampled from all locations and receptacles, at the category-level. If there are multiple receptacles of the  category, we use a second-stage uniform sampling to assign the object to one of the receptacle instance.
\begin{table}[tp!]
\centering\small
\vspace{-0.5em}
\setlength{\tabcolsep}{5pt}
\begin{tabular}{ll}
\toprule
        \textbf{Category} & \textbf{Valid Positions} \\ \midrule
        
furniture
&floor \\

vessel				&countertop table cabinet
\\
tableware			&countertop table cabinet dishwasher refrigerator sink
\\
utensil				&countertop table cabinet dishwasher refrigerator sink
\\
bag					&floor countertop table sofa bed
\\
bucket				&floor countertop table
\\
tray 				&countertop table cabinet refrigerator
\\
basket				&floor countertop table shelf cabinet sofa bed
\\
box					&floor countertop table shelf cabinet sofa bed
\\
package				&floor countertop table shelf cabinet sofa bed
\\
ashcan				&floor
\\
xmas tree			&floor
\\
xmas stocking		&floor countertop table shelf cabinet sofa bed
\\\midrule

fruit 				&table countertop refrigerator						utensil
\\
vegetable			&table countertop refrigerator stove					utensil
\\
drink				&table countertop refrigerator cabinet   			bag
\\
protein				&table countertop refrigerator stove					utensil
\\
flavorer			&table countertop refrigerator cabinet   			bag
\\
baked food			&table countertop refrigerator oven					tray
\\
snack				&table countertop refrigerator microwave				tray
\\
prepared food		&table countertop refrigerator microwave				tray
\\\midrule

metal tool			&countertop table cabinet shelf						furniture
\\
electric equipment	&countertop table cabinet shelf                      furniture
\\
electrical device	&countertop table cabinet shelf						furniture
\\
toiletry			&cabinet toilet bathtub sink pool 					bag
\\
writing tool		&countertop table cabinet shelf						bag
\\
piece of cloth		&cabinet toilet bathtub sink pool					bucket
\\
cleaning tool		&cabinet toilet bathtub sink pool 					bucket
\\
cleansing			&cabinet toilet bathtub sink pool 					bucket
\\
cutlery				&countertop table cabinet dishwasher refrigerator	utensil
\\
illumination tool	&countertop table sofa bed shelf
\\\midrule

decoration			&cabinet sofa bed 									package
\\
paper product		&cabinet sofa bed 									package
\\
footwear			&cabinet floor
\\
headwear			&cabinet sofa bed 									package
\\
clothing			&cabinet sofa bed 									package
\\
building materials	&pool
\\
plaything			&cabinet sofa bed 									package
\\

\bottomrule
\end{tabular}
\vspace{0.2cm}
\caption{Valid positions for each category when sampling the initial state.}
\label{tab:positions}
\end{table}

The last step is to assign attribute values for each object, based on the following rules:
\begin{itemize}
    \item No objects are open, sliced, or soaked initially. Only the refrigerator is toggled on.
    \item Objects at ovens, stoves, and microwaves are cooked. Objects in refrigerators are frozen.
    \item Objects are stained (dusty) with probability $1/3$ if stainable (dustyable).
    \item The size and color properties of objects are uniformly sampled.
\end{itemize}

\subsection{Goal Space}
\label{sec:data-goal-space}
Since all object categories and states above are inherited from \behavior, we can design the goal space based on human-annotated household tasks in \behavior. 

\xhdr{Compositional goal templates.} We summarize the original 100 tasks (by removing duplicates and incompatible ones) into 69 goal templates, represented in first-order logic statements. All the templates are shown in \tbl{tab:goals} to \tbl{tab:goals5}. To instantiate from a template, we first sample a corresponding concrete object category to replace each subclass in brackets. For example, we can replace ``?[fruit]'' by ``apple'' and ``?[vessel]'' by ``jar.'' As a result, each goal templates can be instantiated into a large number of concrete goals. 
\begin{table}[tp!]
\centering\small
\vspace{-0.5em}
\setlength{\tabcolsep}{5pt}
\begin{tabular}{ll}
\toprule
        \textbf{Goal Name} & \textbf{First-Order Logic Formula} \\ \midrule
\multirow{4}{*}{assembling gift baskets}
& $\forall y, \exists x, \text{?[illumination tool]}(x) \land ( \text{basket}(y)\Rightarrow \text{in}(x,y))$ \\
& $\forall y, \exists x, \text{?[snack]}(x) \land ( \text{basket}(y)\Rightarrow \text{in}(x,y))$ \\
& $\forall y, \exists x, \text{?[baked food]}(x) \land ( \text{basket}(y)\Rightarrow \text{in}(x,y))$ \\
& $\forall y, \exists x, \text{?[decoration]}(x) \land ( \text{basket}(y)\Rightarrow \text{in}(x,y))$ \\\midrule
\multirow{3}{*}{bottling fruit}
& $\exists y_1,y_2, \text{jar}(y_1),\text{jar}(y_2),\ldots$\\
& \qquad$\forall x, \text{?[fruit]}_1(x)\Rightarrow \text{in}(x,y_1)$ \\
& \qquad$\forall x, \text{?[fruit]}_2(x)\Rightarrow \text{in}(x,y_2)$ \\\midrule
boxing books up for storage
& $\exists y, \text{box}(y), \forall x, \text{?[paper product]}(x)\Rightarrow \text{in}(x,y)$ \\
\midrule
bringing in wood
& $\exists y, \text{floor}(y), \forall x, \text{?[building material]}(x)\Rightarrow \text{on}(x,y)$ \\
\midrule
brushing lint off clothing
& $\exists y, \text{bed}(y), \forall x, \text{?[clothing]}(x)\Rightarrow \text{on}(x,y) \land \neg \text{dusty}(x)$ \\ \midrule
\multirow{4}{*}{chopping vegetables}
& $\exists y, \text{?[tableware]}(y), \forall x, \text{?[fruit]}_1(x)\Rightarrow \text{in}(x,y) \land \text{sliced}(x)$ \\
& $\exists y, \text{?[tableware]}(y), \forall x, \text{?[fruit]}_2(x)\Rightarrow \text{in}(x,y) \land \text{sliced}(x)$ \\
& $\exists y, \text{?[tableware]}(y), \forall x, \text{?[vegetable]}_1(x)\Rightarrow \text{in}(x,y) \land \text{sliced}(x)$ \\
& $\exists y, \text{?[tableware]}(y), \forall x, \text{?[vegetable]}_2(x)\Rightarrow \text{in}(x,y) \land \text{sliced}(x)$ \\\midrule
\multirow{5}{*}{cleaning bathrooms}
& $\forall x, \text{toilet}(x) \Rightarrow \neg\text{stained}(x)$\\
& $\forall x, \text{bathtub}(x) \Rightarrow \neg\text{stained}(x)$\\
& $\forall x, \text{sink}(x) \Rightarrow \neg\text{stained}(x)$\\
& $\forall x, \text{floor}(x) \Rightarrow \neg\text{stained}(x)$\\
& $\exists y, \text{bucket}(y), \exists x, \text{rag}(x)\land \text{in}(x,y) \land\text{soaked}(x)$ \\ \midrule

\multirow{4}{*}{cleaning bedrooms}
& $\exists y, \text{cabinet}(y), \forall x, \text{?[clothing]}(x)\Rightarrow \text{in}(x,y)\land \neg \text{dusty}(y)$ \\
& $\exists y, \text{cabinet}(y), \forall x, \text{?[decoration]}(x)\Rightarrow \text{in}(x,y)$ \\
& $\exists y, \text{cabinet}(y), \forall x, \text{?[toiletry]}(x)\Rightarrow \text{in}(x,y)$ \\
& $\exists y, \text{cabinet}(y), \forall x, \text{?[paper product]}(x)\Rightarrow \text{in}(x,y)$ \\
\midrule

\multirow{3}{*}{cleaning closet}
& $\exists y, \text{cabinet}(y), \forall x, \text{?[decoration]}(x)\Rightarrow \text{in}(x,y)\land \neg \text{dusty}(y)$ \\
& $\exists y, \text{shelf}(y), \forall x, \text{?[headwear]}(x)\Rightarrow \text{on}(x,y)\land \neg \text{dusty}(y)$ \\
& $\exists y, \text{floor}(y), \forall x, \text{?[footwear]}(x)\Rightarrow \text{on}(x,y)\land \neg \text{dusty}(y)$ \\
\midrule

cleaning floors
& $\forall x, \text{floor}(x) \Rightarrow \neg\text{stained}(x) \land \neg\text{dusty}(x) $\\\midrule

\multirow{5}{*}{cleaning garage}
& $\forall x, \text{floor}(x) \Rightarrow \neg\text{dusty}(x)$\\
& $\forall x, \text{cabinet}(x) \Rightarrow \neg\text{stained}(x)$\\
& $\forall x, \text{cabinet}(x) \Rightarrow \neg\text{dusty}(x)$ \\
& $\exists y, \text{ashcan}(y), \forall x, \text{?[paper product]}(x)\Rightarrow \text{in}(x,y)$ \\
& $\exists y, \text{table}(y), \forall x, \text{?[vessel]}(x)\Rightarrow \text{on}(x,y)$ \\
\midrule

cleaning high chair
& $\forall x, \text{?[furniture]}(x) \Rightarrow \neg\text{dusty}(x)$\\\midrule

\multirow{3}{*}{cleaning kitchen cupboard}
& $\forall x, \text{cabinet}(x) \Rightarrow \neg\text{dusty}(x)$ \\
& $\exists y, \text{cabinet}(y), \forall x, \text{?[tableware]}_1(x)\Rightarrow \text{in}(x,y)$ \\
& $\exists y, \text{cabinet}(y), \forall x, \text{?[tableware]}_2(x)\Rightarrow \text{in}(x,y)$ \\\midrule

cleaning microwave oven
& $\forall x, \text{microwave}(x) \Rightarrow \neg\text{stained}(x) \land \neg\text{dusty}(x)$\\\midrule

\multirow{3}{*}{cleaning out drawers}
& $\exists y, \text{sink}(y), \forall x, \text{?[piece of cloth]}(x)\Rightarrow \text{in}(x,y)$ \\
& $\exists y, \text{sink}(y), \forall x, \text{?[tableware]}(x)\Rightarrow \text{in}(x,y)$ \\
& $\exists y, \text{sink}(y), \forall x, \text{?[cutlery]}(x)\Rightarrow \text{in}(x,y)$ \\
\midrule

\multirow{2}{*}{cleaning oven}
& $\forall x, \text{oven}(x) \Rightarrow \neg\text{stained}(x)$\\
& $\exists x, \text{rag}(x) \land \text{soaked}(x)$
\\\midrule

\multirow{2}{*}{cleaning shoes}
& $\exists y, \text{floor}(y), \forall x, \text{?[footwear]}(x)\Rightarrow \text{on}(x,y)$ \\
& $\exists y, \text{floor}(y), \forall x, \text{?[footwear]}(x)\Rightarrow \text{on}(x,y)\land \neg \text{dusty}(x)$ \\
\midrule

\multirow{3}{*}{cleaning stove}
& $\forall x, \text{stove}(x) \Rightarrow \neg\text{stained}(x)\land \neg\text{dusty}(x)$\\
& $\forall x, \text{rag}(x) \Rightarrow\text{soaked}(x)$\\
& $\forall x, \text{dishtowel}(x) \Rightarrow\text{soaked}(x)$
\\\midrule

cleaning table after clearing
& $\forall x, \text{table}(x) \Rightarrow \neg\text{stained}(x)$\\\midrule

\multirow{3}{*}{cleaning the pool}
& $\forall x, \text{pool}(x) \Rightarrow \neg\text{stained}(x)$\\
& $\exists y, \text{shelf}(y), \forall x, \text{scrub brush}(x)\Rightarrow \text{on}(x,y)$ \\
& $\exists y, \text{floor}(y), \forall x, \text{cleansing}(x)\Rightarrow \text{on}(x,y)$ 
\\

\bottomrule
\end{tabular}
\vspace{0.2cm}
\caption{The goal names are collected from original BEHAVIOR-100 tasks, and the formulas are the rewritten version for original goal predicates.}
\label{tab:goals}
\end{table}

\begin{table}[tp!]
\centering\small
\vspace{-0.5em}
\setlength{\tabcolsep}{5pt}
\begin{tabular}{ll}
\toprule
        \textbf{Goal Name} & \textbf{First-Order Logic Formula} \\ 
        \midrule
\multirow{3}{*}{cleaning toilet}
& $\forall x, \text{toilet}(x) \Rightarrow \neg\text{stained}(x)$\\
& $\exists y, \text{floor}(y), \forall x, \text{scrub brush}(x)\Rightarrow \text{on}(x,y)$ \\
& $\exists y, \text{floor}(y), \forall x, \text{cleansing}(x)\Rightarrow \text{on}(x,y)$ 
\\\midrule

\multirow{7}{*}{cleaning up after a meal}
& $\forall x, \text{table}(x) \Rightarrow \neg\text{stained}(x)$\\
& $\forall x, \text{floor}(x) \Rightarrow \neg\text{stained}(x)$\\
& $\forall x, \text{?[furniture]}(x) \Rightarrow \neg\text{dusty}(x)$\\
& $\forall x, \text{?[tableware]}_1(x) \Rightarrow \neg\text{dusty}(x)$\\
& $\forall x, \text{?[tableware]}_2(x) \Rightarrow \neg\text{dusty}(x)$\\
& $\forall x, \text{?[tableware]}_3(x) \Rightarrow \neg\text{dusty}(x)$\\
& $\exists y, \text{?[bag]}(y), \forall x, \text{?[snack]}(x)\Rightarrow \text{in}(x,y)$
\\\midrule

\multirow{4}{*}{cleaning up refrigerator}
& $\exists y, \text{sink}(y), \forall x, \text{rag}(x)\Rightarrow \text{in}(x,y)$ \\
& $\exists y, \text{sink}(y), \forall x, \text{cleansing}(x)\Rightarrow \text{in}(x,y)$ \\
& $\exists y, \text{refrigerator}(y), \forall x, \text{tray}(x)\Rightarrow \text{in}(x,y)\land \neg \text{dusty}(x)\land \neg \text{stained}(y)$ \\
& $\exists y, \text{sink}(y), \forall x, \text{?[tableware]}(x)\Rightarrow \text{in}(x,y)\land \neg \text{dusty}(x)$ \\
\midrule

\multirow{7}{*}{cleaning up the kitchen only}
& $\exists y, \text{countertop}(y), \forall x, \text{?[vessel]}(x)\Rightarrow \text{on}(x,y)$ \\
& $\exists y, \text{sink}(y), \forall x, \text{?[cleansing]}(x)\Rightarrow \text{in}(x,y)$ \\
& $\exists y, \text{cabinet}(y), \forall x, \text{?[flavorer]}(x)\Rightarrow \text{in}(x,y)$ \\
& $\exists y, \text{cabinet}(y), \forall x, \text{?[tableware]}(x)\Rightarrow \text{in}(x,y)\land \neg \text{dusty}(x)\land \neg \text{dusty}(y)$ \\
& $\exists y, \text{sink}(y), \forall x, \text{rag}(x)\Rightarrow \text{in}(x,y)$ \\
& $\exists y, \text{refrigerator}(y), \forall x, \text{?[utensil]}(x)\Rightarrow \text{in}(x,y)$ \\
& $\exists y, \text{refrigerator}(y), \forall x, \text{?[fruit]}(x)\Rightarrow \text{in}(x,y)$ \\
\midrule

\multirow{3}{*}{clearing the table after dinner}
& $\exists y_1, \text{bucket}(y_1), \forall x, \text{?[cutlery]}_1(x)\Rightarrow \text{in}(x,y_1)$ \\
& $\exists y_2, \text{bucket}(y_2), \forall x, \text{?[cutlery]}_2(x)\Rightarrow \text{in}(x,y_2)$ \\
& $\exists y, \text{bucket}(y), \forall x, \text{?[flavorer]}(x)\Rightarrow \text{in}(x,y)$ \\
\midrule

\multirow{3}{*}{collect misplaced items}
& $\exists y, \text{table}(y), \forall x, \text{?[footwear]}(x)\Rightarrow \text{on}(x,y)$ \\
& $\exists y, \text{table}(y), \forall x, \text{?[decoration]}(x)\Rightarrow \text{on}(x,y)$ \\
& $\exists y, \text{table}(y), \forall x, \text{?[paper product]}(x)\Rightarrow \text{on}(x,y)$ \\
\midrule

collecting aluminum cans
& $\exists y, \text{ashcan}(y), \forall x, \text{?[drink]}(x)\Rightarrow \text{in}(x,y)$ \\

\multirow{3}{*}{filling a Christmas stocking}
& $\forall y, \exists x, \text{?[plaything]}(x) \land ( \text{Xmas stocking}(y)\Rightarrow \text{in}(x,y))$ \\
& $\forall y, \exists x, \text{?[snack]}(x) \land ( \text{Xmas stocking}(y)\Rightarrow \text{in}(x,y))$ \\
& $\forall y, \exists x, \text{?[writing tool]}(x) \land ( \text{Xmas stocking}(y)\Rightarrow \text{in}(x,y))$ \\
\midrule

\multirow{6}{*}{filling a Easter basket}
& $\exists y, \text{countertop}(y), \forall x, \text{?[basket]}(x)\Rightarrow \text{on}(x,y)$ \\
& $\forall y, \exists x, \text{?[protein]}(x) \land ( \text{basket}(y)\Rightarrow \text{in}(x,y))$ \\
& $\forall y, \exists x, \text{?[snack]}(x) \land ( \text{basket}(y)\Rightarrow \text{in}(x,y))$ \\
& $\forall y, \exists x, \text{?[paper product]}(x) \land ( \text{basket}(y)\Rightarrow \text{in}(x,y))$ \\
& $\forall y, \exists x, \text{?[plaything]}(x) \land ( \text{basket}(y)\Rightarrow \text{in}(x,y))$ \\
& $\exists y, \text{basket}(y), \forall x, \text{?[decoration]}(x)\Rightarrow \text{in}(x,y)$ \\
\midrule

installing a fax machine
& $\exists y, \text{table}(y), \forall x, \text{?[electric equipment]}(x)\Rightarrow \text{on}(x,y)) \land \text{toggled}(x)$ \\
\midrule
\multirow{3}{*}{installing alarms}
& $\forall y, \exists x, \text{?[electrical device]}(x) \land ( \text{table}(y)\Rightarrow \text{on}(x,y))$ \\
& $\forall y, \exists x, \text{?[electrical device]}(x) \land ( \text{countertop}(y)\Rightarrow \text{on}(x,y))$ \\
& $\forall y, \exists x, \text{?[electrical device]}(x) \land ( \text{sofa}(y)\Rightarrow \text{on}(x,y))$ \\
\midrule

laying tile floors
& $\exists y, \text{floor}(y), \forall x, \text{?[building material]}(x)\Rightarrow \text{on}(x,y)$ \\
\midrule

\multirow{3}{*}{loading the dishwasher}
& $\exists y, \text{sink}(y), \forall x, \text{?[tableware]}_1(x)\Rightarrow \text{in}(x,y)$ \\
& $\exists y, \text{sink}(y), \forall x, \text{?[tableware]}_2(x)\Rightarrow \text{in}(x,y)$ \\
& $\exists y, \text{sink}(y), \forall x, \text{?[vessel]}(x)\Rightarrow \text{in}(x,y)$ \\
\midrule

moving boxes to storage
& $\exists y, \text{floor}(y), \forall x, \text{box}(x)\Rightarrow \text{on}(x,y)$ \\
\midrule

opening packages
& $\forall x, \text{package}(x)\Rightarrow \text{open}(x)$ \\
\midrule

\multirow{5}{*}{organizing boxes in garage}
& $\exists y, \text{floor}(y), \forall x, \text{box}(x)\Rightarrow \text{on}(x,y)$ \\
& $\exists y_1,y_2,y_3, \text{box}(y_1),\text{box}(y_2),\text{box}(y_3), \ldots$ \\
& \qquad $\forall x, \text{?[plaything]}(x)\Rightarrow \text{in}(x,y_1)$ \\
& \qquad $\forall x, \text{?[cutlery]}(x)\Rightarrow \text{in}(x,y_2)$ \\
& \qquad$\forall x, \text{?[cleansing]}(x)\Rightarrow \text{in}(x,y_3)$ \\

\bottomrule
\end{tabular}
\vspace{0.2cm}
\caption{(Cont'd.) The goal names are collected from original BEHAVIOR-100 tasks, and the formulas are the rewritten version for original goal predicates.}
\label{tab:goals2}
\end{table}

\begin{table}[tp!]
\centering\small
\vspace{-0.5em}
\setlength{\tabcolsep}{5pt}
\begin{tabular}{ll}
\toprule
        \textbf{Goal Name} & \textbf{First-Order Logic Formula} \\ 
        \midrule
\multirow{2}{*}{organizing file cabinet}
& $\exists y, \text{table}(y), \forall x, \text{?[writing tool]}(x)\Rightarrow \text{on}(x,y)$ \\
& $\exists y, \text{cabinet}(y), \forall x, \text{?[paper product]}(x)\Rightarrow \text{in}(x,y)$ \\
\midrule

\multirow{5}{*}{organizing school stuff}
& $\exists y, \text{backpack}(y), \ldots$\\
& \qquad $\exists x, \text{?[paper product]}(x)\land \text{in}(x,y)$ \\
& \qquad $\exists x, \text{?[writing tool]}_1(x)\land \text{in}(x,y)$ \\
& \qquad $\exists x, \text{?[writing tool]}_2(x)\land \text{in}(x,y)$ \\
& \qquad $\exists x, \text{?[electrical device]}(x)\land \text{in}(x,y)$ \\
\midrule

\multirow{5}{*}{packing adult's bag}
& $\exists y, \text{backpack}(y), \ldots$\\
& \qquad $\forall x, \text{?[decoration]}(x)\Rightarrow \text{in}(x,y)$ \\
& \qquad $\exists x, \text{?[toiletry]}_1(x)\land \text{in}(x,y)$ \\
& \qquad $\exists x, \text{?[toiletry]}_2(x)\land \text{in}(x,y)$ \\
& \qquad $\exists x, \text{?[electrical device]}(x)\land \text{in}(x,y)$ \\
\midrule

\multirow{6}{*}{packing bags or suitcase}
& $\exists y, \text{briefcase}(y), \ldots$\\
& \qquad $\forall x, \text{?[clothing]}(x)\Rightarrow \text{in}(x,y)$ \\
& \qquad $\exists x, \text{?[toiletry]}(x)\land \text{in}(x,y)$ \\
& \qquad $\exists x, \text{?[cleansing]}_1(x)\land \text{in}(x,y)$ \\
& \qquad $\exists x, \text{?[cleansing]}_2(x)\land \text{in}(x,y)$ \\
& \qquad $\exists x, \text{?[paper product]}(x)\land \text{in}(x,y)$ \\
\midrule

\multirow{5}{*}{packing boxes for household move or trip}
& $\exists y_1,y_2, \text{box}(y_1), \text{box}(y_2), \ldots$\\
& \qquad $\forall x, \text{?[cutlery]}(x)\Rightarrow \text{in}(x,y_1)$ \\
& \qquad $\exists x, \text{?[piece of cloth]}(x)\land \text{in}(x,y_1)$ \\
& \qquad $\forall x, \text{book}(x)\Rightarrow \text{in}(x,y_2)$ \\
& \qquad $\exists x, \text{?[clothing]}(x)\land \text{in}(x,y_2)$ \\
\midrule

\multirow{6}{*}{packing child's bag}
& $\exists y, \text{backpack}(y), \ldots$\\
& \qquad $\exists x, \text{?[headwear]}(x)\land \text{in}(x,y)$ \\
& \qquad $\exists x, \text{?[decoration]}(x)\land \text{in}(x,y)$ \\
& \qquad $\exists x, \text{?fruit]}(x)\land \text{in}(x,y)$ \\
& \qquad $\exists x, \text{?[paper product]}(x)\land \text{in}(x,y)$ \\
& \qquad $\exists x, \text{?[electrical device]}(x)\land \text{in}(x,y)$ \\
\midrule

\multirow{5}{*}{packing box for work}
& $\exists y, \text{box}(y), \ldots$\\
& \qquad $\exists x, \text{?[fruit]}(x)\land \text{in}(x,y)$ \\
& \qquad $\exists x, \text{?[drink]}(x)\land \text{in}(x,y)$ \\
& \qquad $\exists x, \text{?[snack]}_1(x)\land \text{in}(x,y)$ \\
& \qquad $\exists x, \text{?[snack]}_2(x)\land \text{in}(x,y)$ \\
\midrule

\multirow{11}{*}{packing lunches}
& $\exists y_1,y_2, \text{box}(y_1), \text{box}(y_2), \ldots$\\
& \qquad $\exists x, \text{?[snack]}_1(x)\land \text{in}(x,y_1)$ \\
& \qquad $\exists x, \text{?[snack]}_1(x)\land \text{in}(x,y_2)$ \\
& \qquad $\exists x, \text{?[baked food]}_1(x)\land \text{in}(x,y_1)$ \\
& \qquad $\exists x, \text{?[baked food]}_1(x)\land \text{in}(x,y_2)$ \\
& \qquad $\exists x, \text{?[prepared food]}(x)\land \text{in}(x,y_1)$ \\
& \qquad $\exists x, \text{?[snack]}_2(x)\land \text{in}(x,y_2)$ \\
& \qquad $\exists x, \text{?[drink]}_1(x)\land \text{in}(x,y_1)$ \\
& \qquad $\exists x, \text{?[drink]}_2(x)\land \text{in}(x,y_2)$ \\
& \qquad $\exists x, \text{?[fruit]}_1(x)\land \text{in}(x,y_1)$ \\
& \qquad $\exists x, \text{?[fruit]}_2(x)\land \text{in}(x,y_2)$ \\

\midrule

\multirow{9}{*}{packing picnics}
& $\exists y_1,y_2,y_3, \text{box}(y_1), \text{box}(y_2),\text{box}(y_3), \ldots$\\
& \qquad $\exists x, \text{?[snack]}_1(x)\land \text{in}(x,y_1)$ \\
& \qquad $\exists x, \text{?[snack]}_2(x)\land \text{in}(x,y_1)$ \\
& \qquad $\exists x, \text{?[fruit]}_1(x)\land \text{in}(x,y_2)$ \\
& \qquad $\exists x, \text{?[fruit]}_2(x)\land \text{in}(x,y_2)$ \\
& \qquad $\exists x, \text{?[fruit]}_3(x)\land \text{in}(x,y_2)$ \\
& \qquad $\exists x, \text{?[drink]}_1(x)\land \text{in}(x,y_3)$ \\
& \qquad $\exists x, \text{?[drink]}_2(x)\land \text{in}(x,y_3)$ \\
& \qquad $\exists x, \text{?[drink]}_3(x)\land \text{in}(x,y_3)$ \\

\midrule

\multirow{3}{*}{picking up take out food}
& $\exists y, \text{box}(y), \forall x, \text{?[prepared food]}(x)\Rightarrow \text{in}(x,y)$ \\
& $\exists y, \text{box}(y), \forall x, \text{?[snack]}(x)\Rightarrow \text{in}(x,y)$ \\
& $\exists y, \text{floor}(y), \forall x, \text{?[box]}(x)\Rightarrow \text{on}(x,y) \land \text{open}(x)$ \\

\bottomrule
\end{tabular}
\vspace{0.2cm}
\caption{(Cont'd.) The goal names are collected from original BEHAVIOR-100 tasks, and the formulas are the rewritten version for original goal predicates.}
\label{tab:goals3}
\end{table}

\begin{table}[tp!]
\centering\small
\vspace{-0.5em}
\setlength{\tabcolsep}{5pt}
\begin{tabular}{ll}
\toprule
        \textbf{Goal Name} & \textbf{First-Order Logic Formula} \\ 
        \midrule

\multirow{2}{*}{picking up trash}
& $\exists y, \text{ashcan}(y), \forall x, \text{?[paper product]}(x)\Rightarrow \text{in}(x,y)$ \\
& $\exists y, \text{ashcan}(y), \forall x, \text{?[drink]}(x)\Rightarrow \text{in}(x,y)$ \\
\midrule

\multirow{2}{*}{polishing furniture}
& $\forall x, \text{table}(x)\Rightarrow \neg\text{dusty}(x)$ \\
& $\forall x, \text{shelf}(x)\Rightarrow \neg\text{dusty}(x)$ \\
\midrule

\multirow{2}{*}{polishing shoes}
& $\exists y, \text{sink}(y), \forall x, \text{rag}(x)\Rightarrow \text{in}(x,y) \land \text{soaked}(x)$ \\
& $\forall x, \text{?[footwear]}(x)\Rightarrow \neg\text{dusty}(x)$ \\
\midrule

\multirow{2}{*}{polishing silver}
& $\exists y, \text{cabinet}(y), \forall x, \text{?[cutlery]}(x)\Rightarrow \text{in}(x,y) \land \neg \text{dusty}(x)$ \\
& $\exists y, \text{cabinet}(y), \forall x, \text{rag}(x)\Rightarrow \text{in}(x,y)$ \\
\midrule

\multirow{9}{*}{preparing food}
& $\exists y_1,y_2, \text{plate}(y_1), \text{plate}(y_2), \ldots$\\
& \qquad $\exists x, \text{?[vegetable]}_1(x)\Rightarrow \text{in}(x,y_1)$ \\
& \qquad $\exists x, \text{?[vegetable]}_1(x)\Rightarrow \text{in}(x,y_2)$ \\
& \qquad $\exists x, \text{?[vegetable]}_2(x)\Rightarrow \text{in}(x,y_1)$ \\
& \qquad $\exists x, \text{?[vegetable]}_2(x)\Rightarrow \text{in}(x,y_2)$ \\
& \qquad $\exists x, \text{?[vegetable]}_3(x)\Rightarrow \text{in}(x,y_1)\land \text{sliced}(x)$ \\
& \qquad $\exists x, \text{?[vegetable]}_3(x)\Rightarrow \text{in}(x,y_2)\land \text{sliced}(x)$ \\
& \qquad $\exists x, \text{?[fruit]}_1(x)\Rightarrow \text{in}(x,y_1)\land \text{sliced}(x)$ \\
& \qquad $\exists x, \text{?[fruit]}_1(x)\Rightarrow \text{in}(x,y_2)\land \text{sliced}(x)$ \\
\midrule

\multirow{2}{*}{preserving food}
& $\exists y, \text{?[vessel]}(y), \forall x, \text{?[fruit]}(x)\Rightarrow \text{in}(x,y) \land \text{sliced}(x)\land \text{cooked}(x)$ \\
& $\exists y, \text{refrigerator}(y), \forall x, \text{?[protein]}(x)\Rightarrow \text{in}(x,y)$ \\
\midrule

\multirow{3}{*}{putting away Christmas decorations}
& $\exists y, \text{cabinet}(y), \forall x, \text{?[decoration]}_1(x)\Rightarrow \text{in}(x,y)$ \\
& $\exists y, \text{cabinet}(y), \forall x, \text{?[decoration]}_2(x)\Rightarrow \text{in}(x,y)$ \\
& $\exists y, \text{cabinet}(y), \forall x, \text{?[decoration]}_3(x)\Rightarrow \text{in}(x,y)$ \\
\midrule

\multirow{3}{*}{putting away Halloween decorations}
& $\exists y, \text{cabinet}(y), \forall x, \text{?[vegetable]}(x)\Rightarrow \text{in}(x,y)$ \\
& $\exists y, \text{cabinet}(y), \forall x, \text{?[illumination tool]}(x)\Rightarrow \text{in}(x,y)$ \\
& $\exists y, \text{table}(y), \forall x, \text{?[vessel]}(x)\Rightarrow \text{on}(x,y)$ \\
\midrule

putting away toys
& $\exists y, \text{box}(y), \forall x, \text{?[plaything]}(x)\Rightarrow \text{in}(x,y)\land \neg \text{open}(y)$ \\
\midrule

putting dishes away after cleaning
& $\exists y, \text{cabinet}(y), \forall x, \text{?[tableware]}(x)\Rightarrow \text{in}(x,y)$ \\
\midrule

\multirow{4}{*}{putting up Christmas decorations inside}
& $\exists y, \text{table}(y), \forall x, \text{?[illumination tool]}(x)\Rightarrow \text{on}(x,y)$ \\
& $\exists y, \text{table}(y), \forall x, \text{?[decoration]}_1(x)\Rightarrow \text{on}(x,y)$ \\
& $\exists y, \text{table}(y), \forall x, \text{?[decoration]}_2(x)\Rightarrow \text{on}(x,y)$ \\
& $\exists y, \text{table}(y), \forall x, \text{?[decoration]}_3(x)\Rightarrow \text{on}(x,y)$ \\
\midrule

re-shelving library books
& $\exists y, \text{shelf}(y), \forall x, \text{?[paper product]}(x)\Rightarrow \text{on}(x,y)$ \\
\midrule

\multirow{12}{*}{serving a meal}
& $\exists y_1,y_2,y_3, \text{table}(y_1), \text{dish}(y_2),\text{dish}(y_3), \ldots$\\
& \qquad $\forall x, \text{dish}(x)\land \text{on}(x,y_1)$ \\
& \qquad $\forall x, \text{bowl}(x)\land \text{on}(x,y_1)$ \\
& \qquad $\forall x, \text{?[cutlery]}_1(x)\land \text{on}(x,y_1)$ \\
& \qquad $\forall x, \text{?[cutlery]}_2(x)\land \text{on}(x,y_1)$ \\
& \qquad $\forall x, \text{?[drink]}(x)\land \text{on}(x,y_1)$ \\
& \qquad $\exists x, \text{?[protein]}(x)\land \text{in}(x,y_2)$ \\
& \qquad $\exists x, \text{?[protein]}(x)\land \text{in}(x,y_2)$ \\
& \qquad $\exists x, \text{?[prepared food]}(x)\land \text{in}(x,y_2)$ \\
& \qquad $\exists x, \text{?[prepared food]}(x)\land \text{in}(x,y_3)$ \\
& \qquad $\exists x, \text{?[baked food]}(x)\land \text{in}(x,y_2)$ \\
& \qquad $\exists x, \text{?[baked food]}(x)\land \text{in}(x,y_3)$ \\

\midrule

\multirow{4}{*}{serving hors d'oeuvres}
& $\exists y, \text{table}(y), \forall x, \text{tray}(x)\Rightarrow \text{on}(x,y)$ \\
& $\exists y, \text{table}(y), \forall x, \text{?[baked food]}(x)\Rightarrow \text{on}(x,y)$ \\
& $\exists y, \text{table}(y), \forall x, \text{?[vegetable]}(x)\Rightarrow \text{on}(x,y)$ \\
& $\exists y, \text{table}(y), \forall x, \text{?[prepared food]}(x)\Rightarrow \text{on}(x,y)$ \\
\midrule

setting up candles
& $\forall y, \exists x, \text{?[illumination tool]}(x) \land (\text{?[furniture]}(y)\Rightarrow \text{on}(x,y))$ \\
\midrule

\multirow{2}{*}{sorting books}
& $\exists y, \text{shelf}(y), \forall x, \text{?[paper product]}_1(x)\Rightarrow \text{on}(x,y)$ \\
& $\exists y, \text{shelf}(y), \forall x, \text{?[paper product]}_2(x)\Rightarrow \text{on}(x,y)$ \\

\bottomrule
\end{tabular}
\vspace{0.2cm}
\caption{(Cont'd.) The goal names are collected from original BEHAVIOR-100 tasks, and the formulas are the rewritten version for original goal predicates.}
\label{tab:goals4}
\end{table}

\begin{table}[tp!]
\centering\small
\vspace{-0.5em}
\setlength{\tabcolsep}{5pt}
\begin{tabular}{ll}
\toprule
        \textbf{Goal Name} & \textbf{First-Order Logic Formula} \\ 
        \midrule

\multirow{4}{*}{storing food}
& $\exists y, \text{cabinet}(y), \forall x, \text{?[prepared food]}(x)\Rightarrow \text{in}(x,y)$ \\
& $\exists y, \text{cabinet}(y), \forall x, \text{?[snack]}(x)\Rightarrow \text{in}(x,y)$ \\
& $\exists y, \text{cabinet}(y), \forall x, \text{?[flavorer]}_1(x)\Rightarrow \text{in}(x,y)$ \\
& $\exists y, \text{cabinet}(y), \forall x, \text{?[flavorer]}_2(x)\Rightarrow \text{in}(x,y)$ \\
\midrule

\multirow{4}{*}{storing the groceries}
& $\exists y, \text{refrigerator}(y), \forall x, \text{?[fruit]}(x)\Rightarrow \text{in}(x,y)$ \\
& $\exists y, \text{refrigerator}(y), \forall x, \text{?[protein]}(x)\Rightarrow \text{in}(x,y)$ \\
& $\exists y, \text{refrigerator}(y), \forall x, \text{?[vegetable]}_1(x)\Rightarrow \text{in}(x,y)$ \\
& $\exists y, \text{refrigerator}(y), \forall x, \text{?[vegetable]}_2(x)\Rightarrow \text{in}(x,y)$ \\
\midrule

\multirow{3}{*}{thawing frozen food}
& $\exists y, \text{sink}(y), \forall x, \text{?[fruit]}(x)\Rightarrow \text{in}(x,y)$ \\
& $\exists y, \text{sink}(y), \forall x, \text{?[protein]}(x)\Rightarrow \text{in}(x,y)$ \\
& $\exists y, \text{sink}(y), \forall x, \text{?[vegetable]}(x)\Rightarrow \text{in}(x,y)$ \\
\midrule

throwing away leftovers
& $\exists y, \text{ashcan}(y), \forall x, \text{?[snack]}(x)\Rightarrow \text{in}(x,y)$\\
\midrule

\multirow{3}{*}{washing dishes}
& $\forall x, \text{?[tableware]}_1(x)\Rightarrow \neg \text{dusty}(x)$ \\
& $\forall x, \text{?[tableware]}_2(x)\Rightarrow \neg \text{dusty}(x)$ \\
& $\forall x, \text{?[tableware]}_3(x)\Rightarrow \neg \text{dusty}(x)$ \\
\midrule

\multirow{3}{*}{washing pots and pans}
& $\exists y, \text{cabinet}(y), \forall x, \text{?[utensil}(x)\Rightarrow \text{in}(x,y)\land \neg \text{dusty}(x)$ \\
& $\exists y, \text{cabinet}(y), \forall x, \text{?[vessel]}_1(x)\Rightarrow \text{in}(x,y)\land \neg \text{dusty}(x)$ \\
& $\exists y, \text{cabinet}(y), \forall x, \text{?[vessel]}_2(x)\Rightarrow \text{in}(x,y)\land \neg \text{dusty}(x)$ \\

\bottomrule
\end{tabular}
\vspace{0.2cm}
\caption{(Cont'd.) The goal names are collected from original BEHAVIOR-100 tasks, and the formulas are the rewritten version for original goal predicates.}
\label{tab:goals5}
\end{table}

\xhdr{PDDL setting.} We use PDDL (Planning Domain Definition Language) to set up the \benchmark environment for both human and robot agent. All attributes and relations are represented by predicates with one or two arguments. All the actions are specified with particular preconditions (\ie, constraints) and effects (\ie, semantics). The goal state for each task is represented as the first-order logic statements defined above.

\xhdr{Version 2.} We choose a smaller subset of 25 goal templates for Version 2 dataset, listed as followings: boxing books up for storage, bringing in wood, clearing the table after dinner, collect misplaced items, collecting aluminum cans, installing alarms, laying tile floors, loading the dishwasher, moving boxes to storage, organizing boxes in garage, organizing file cabinet, picking up trash, putting away Christmas decorations, putting away Halloween decorations, putting away toys, putting dishes away after cleaning, putting leftovers away, putting up Christmas decorations inside, re-shelving library books, serving hors d’oeuvres, sorting books, storing food, storing the groceries, thawing frozen food, throwing away leftovers.

\subsection{Quests}
We translate the sampled utterance $u$ natural language (English) following the templates in \tbl{tab:quests}. Specifically, in the table we have a ``Description'' function that takes in a set of specifiers and outputs natural language description. The template for this function consists of four parts: static attributes, non-static attributes, object category and current position. These parts are connected coherently. Usually a description in the quest only specifies some of the four parts. Example descriptions are: ``the large red box on the table,'' ``the uncooked food,'' and ``the sliced one.'' The pronoun ``one'' is used if the specifier does not contain any category information.

\begin{table}[tp!]
\centering\small
\vspace{-0.5em}
\setlength{\tabcolsep}{5pt}
\begin{tabular}{ll}
\toprule
        \textbf{Utterance} & \textbf{Natural Language Instruction} \\ \midrule
bring-me(None) & Bring/Hand/Give me that. \\
bring-me(specifiers) & Bring/Hand/Give me [Description(specifiers)].\\ \midrule
move-to(None, None) & Put it over there.\\
move-to(specifiers, None) & Put [Description(specifiers)] over there.\\
move-to(None, position) & Put/Move it to the [position]. \\
move-to(specifiers, position) & Put/Move [Description(specifiers)] to the [position].\\ \midrule
change-state(None, verb, prep) & [Verb] it [prep].\\
change-state(specifiers, verb, prep) & [Verb] [prep] [Description(specifiers)].\\

\bottomrule
\end{tabular}
\vspace{0.2cm}
\caption{Translation from utterance to natural language instruction. ``Description(specifiers)'' transfers a set of specifiers to a natural language description. The brackets should be replaced by corresponding values, and the words in parentheses should be removed if that attribute doesn't appear in the specifiers. Moreover, we randomly append ``please,'' ``can you'' or nothing to the beginning of an instruction.}
\label{tab:quests}
\end{table}

\subsection{Cost Functions}
To compute the utility of the human and to evaluate the performance of agents, we define the following cost functions.

\xhdr{Human's action cost.} For simplicity, we assume all human actions are unit-cost. Thus, we use the length of action list as the total cost of a trajectory. 

\xhdr{Human's language cost.} We use the following cost function to evaluate the complexity of a quest (both meaning and utterance), which is intuitively the number of specifiers in the statement. The specifiers include: 1) category, 2) attributes and relations, 3) target position (only for ``move-to''-typed quests). Each specifier in the latter two kinds has cost 1. For example, the specifier``large'' costs $1$ and ``cooked, on-table'' costs $2$. For category information, each quest can contain at most one specifier of \{class, subclass, category\}. They cost 1, 2, and 3, respectively. We present more examples for this cost function in \tbl{tab:costs}.

\xhdr{Robot's action cost.} Similar to human's action costs, we assume all robot actions are unit-costs. There are two exceptions: ``examine'' and ``inventory.'' Action ``examine'' gathers the information of objects at the current position. Action ``inventory'' gathers the information of the current holding object. The costs of these two actions are zero: the agent can obtain such information at no cost and at any time. 

\begin{table}[tp!]
\centering\small
\vspace{-0.5em}
\setlength{\tabcolsep}{5pt}
\begin{tabular}{ll}
\toprule
        \textbf{Quest} & \textbf{Cost} \\ \midrule
Bring me that.
& 0\\
Bring me the \textit{food}. 
& 1\\
Bring me the \textit{apple} on the \textit{table}.
& 3+1\\
\midrule
Put it over there.
& 0\\
Put the \textit{large} \textit{receptacle} over there.
& 1+1\\
Move the \textit{large}, \textit{red} \textit{box} to the \textit{sofa}.
& 1+1+2+1\\
\midrule
Slice it up.
& 0\\
Slice up the one on the \textit{table}.
& 1\\
Slice up the \textit{unfrozen} \textit{apple} on the \textit{table}.
& 1+3+1\\
\bottomrule
\end{tabular}
\vspace{0.2cm}
\caption{Examples for quest costs. For easier understanding, we represent the quests in natural language. Note that when we are generating data, we separately process three types of quests, so we only need to compare the costs within a same quest type.}
\label{tab:costs}
\end{table}

\subsection{Hyperparameters}
In our data generation process, we use the following hyperparameters.
\begin{itemize}
    \item When generating the meaning $m$, we need two inverse temperature constants for the Boltzmann distribution: $\beta_1=3, \beta_2=1.5$.
    \item \revise{When generating the utterance $u$, we need three constants for the RSA model: $\alpha=2, \alpha'=1$ and $k = 10$.}
    \item When evaluating the hardness level, we need to estimate the most probable meaning following the RSA model. In this paper, we use same values of $\alpha$ and $k$ as in generation process.

\end{itemize}
\revise{The hyperparameter $\beta_1,\beta_2,\alpha,\alpha'$ in generating meaning and utterance are chosen to balance the generated data. Intuitively, these values describe the habit in asking for help and language use of a particular human. For example, when $\alpha'$ gets larger, the human tends to give use shorter phrase in utterance (like "hand me that"); otherwise, the human tends to use more detailed description in utterance. We choose the values so that different lengths of meaning and utterance can be generated, ensuring the diversity of data.}

\revise{The hyperparameter $k=10$ in RSA model is not sensitive to our generation process. It is chosen to be non-trivial while easy to compute. This parameter is used to compute the distribution for utterance generation, and $k=10$ is already close to the effect of $k\rightarrow \infty$.}

\subsection{Textual Interface}
As claimed in the main text, we construct a textual interface for \benchmark based on a gym-like environment. In this subsection, we present the templates for all valid commands and corresponding observations. The action and observation templates are listed in \tbl{tab:interface}.
\begin{table}[tp!]
\centering\small
\vspace{-0.5em}
\setlength{\tabcolsep}{5pt}
\begin{tabular}{ll}
\toprule
        \textbf{Action} & \textbf{Observation} \\ \midrule

\multirow{3}{*}{(Welcome Message)} & Welcome to the world! \\
& In the room there is a countertop, a sofa, \ldots, a pool. \\
& Now you are standing on the floor. \\ \midrule
\multirow{5}{*}{(Previous Trajectory)} & The human agent has taken a list of actions towards a goal, \\
& which includes: \\
& Human moves to the [location].\\
& Human opens [receptacle] at the [location].\\
& Human picks up [object] at the [location]\ldots\\\midrule
\multirow{2}{*}{(Human's utterance)}
& Human stops and says, ``[Natural Language Instruction].''\\
& Now it is your turn to help human to achieve the goal!\\ \midrule
\multirow{4}{*}{Examine/Look}
& You are at the [location]. You see the [state] [location].\\
& In/On [location] you can see [state] [object] [ID], \ldots\\
& In/On [receptacle] you can see [state] [object] [ID], \ldots \\
& \quad (First the [location] and then the [receptacle] here.)\\ \midrule
\multirow{2}{*}{Inventory}
& You are holding nothing / [state] [object] [ID].\\
& Recall your task: [Previous Trajectory], [Human's Utterance].\\ \midrule

Move to [location]
& You move from [location] to [location].\\
Pick up [movable]
&You pick up the [movable] from [location].\\
Pick up [movable] from [receptacle]
&You pick up the [movable] from the [receptacle] at the [location].\\
put [movable] into/onto [location]
&You put the [movable] into/onto the [location].\\
put [movable] into/onto [receptacle]
&You put the [movable] into/onto the [receptacle] at the [location].\\
take [movable] from human
&You take the [movable] from [location].\\
give [movable] to human
&You give the [movable] to [location].\\
open/close [receptacle/location]
&You open/close the [receptacle/location].\\
toggle on/off [toggleable]
&You toggle the [toggleable] on/off.\\
heat [cookable]
&You heat the [cookable] up with the [location].\\
cool [freezable]
&You cool the [freezable] down with the [location].\\
soak [soakable]
&You make the [soakable] soaked with the [location].\\
slice [sliceable] with [tool]
&You slice up the [sliceable] with the [tool].\\
clean [cleanable] with [tool]
&You clean up the [cleanable] with the [tool].\\

\bottomrule
\end{tabular}
\vspace{0.2cm}
\caption{This table presents the natural language templates for actions and corresponding observations in textual interface. Note that the initial observation is not related to player's command, including welcome message and task description.}
\label{tab:interface}
\end{table}

We have a set of commands for player to move, examine and manipulate objects in the world. The objects in a command should be specified through object identifiers instead of its category name, \ie, they should include IDs.

There is an initial observation when the game starts. It contains the welcome message, previous human actions, examine result at current location, and human's quest statement. Each time when the player enters a command, a new observation is generated, which includes the message of the command's effect. If the command is not valid, the observation will be ``You can't do that'' or ``I can't understand.'' To reduce the processing cost of neural-network-based agents, we have shorten textual observations, especially for the ``examine'' results.
\section{Dataset Statistics}
\label{sec:statistics}

\paragraph{Scene.}

\begin{figure}[tp]
  \centering\small
  \includegraphics[width=\textwidth]{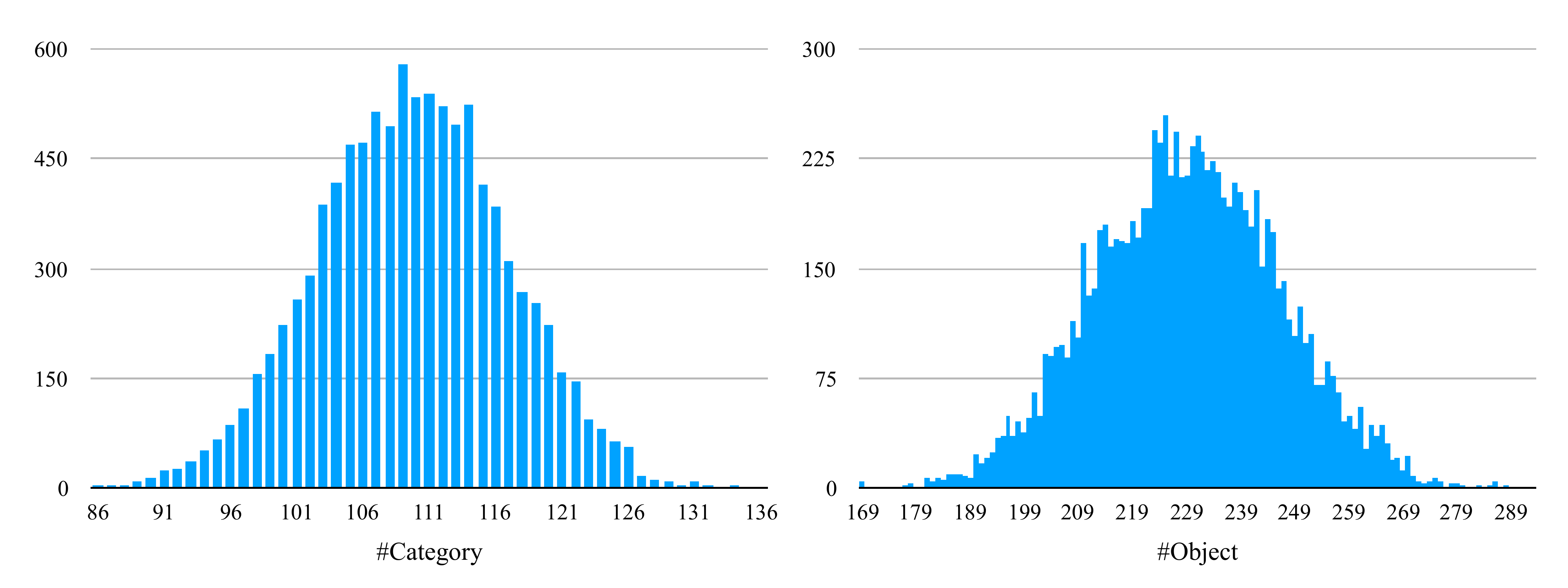}
  \vspace{-1em}
  \caption{Number of categories and objects in the scene.}
  \label{fig:objects}
  \vspace{-1em}
\end{figure}

\fig{fig:objects} shows the number of categories and objects in the sampled scenes. There are 230 objects and 110 categories on average in each scene.

\paragraph{Human actions.}

\begin{figure}[tp]
  \centering\small
  \includegraphics[width=\textwidth]{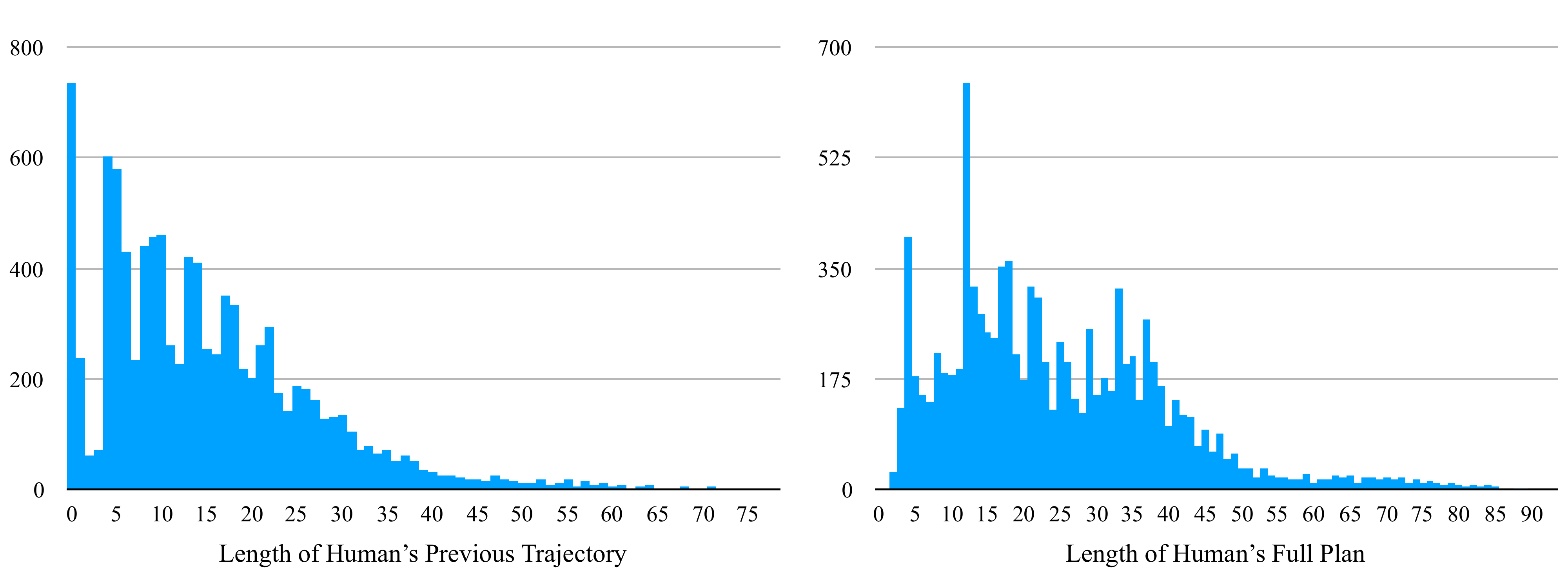}
  \vspace{-1em}
  \caption{Lengths of human's previous trajectory and original full plan.}
  \label{fig:traj}
  \vspace{-1em}
\end{figure}

\fig{fig:traj}(left) shows the length distribution of human's action list before she asks for help, with an average length 15.
\fig{fig:traj}(right) shows the length distribution of human's original full plan towards her goal, with an average length 25.

\paragraph{Demonstration.}
For each episode, we provide an expert demonstration that achieves the subgoal. \camera{The trajectories are generated by applying greedy best-first search algorithms with FF heuristic on each episode. }

\begin{figure}[tp]
  \centering\small
  \includegraphics[width=0.7\textwidth]{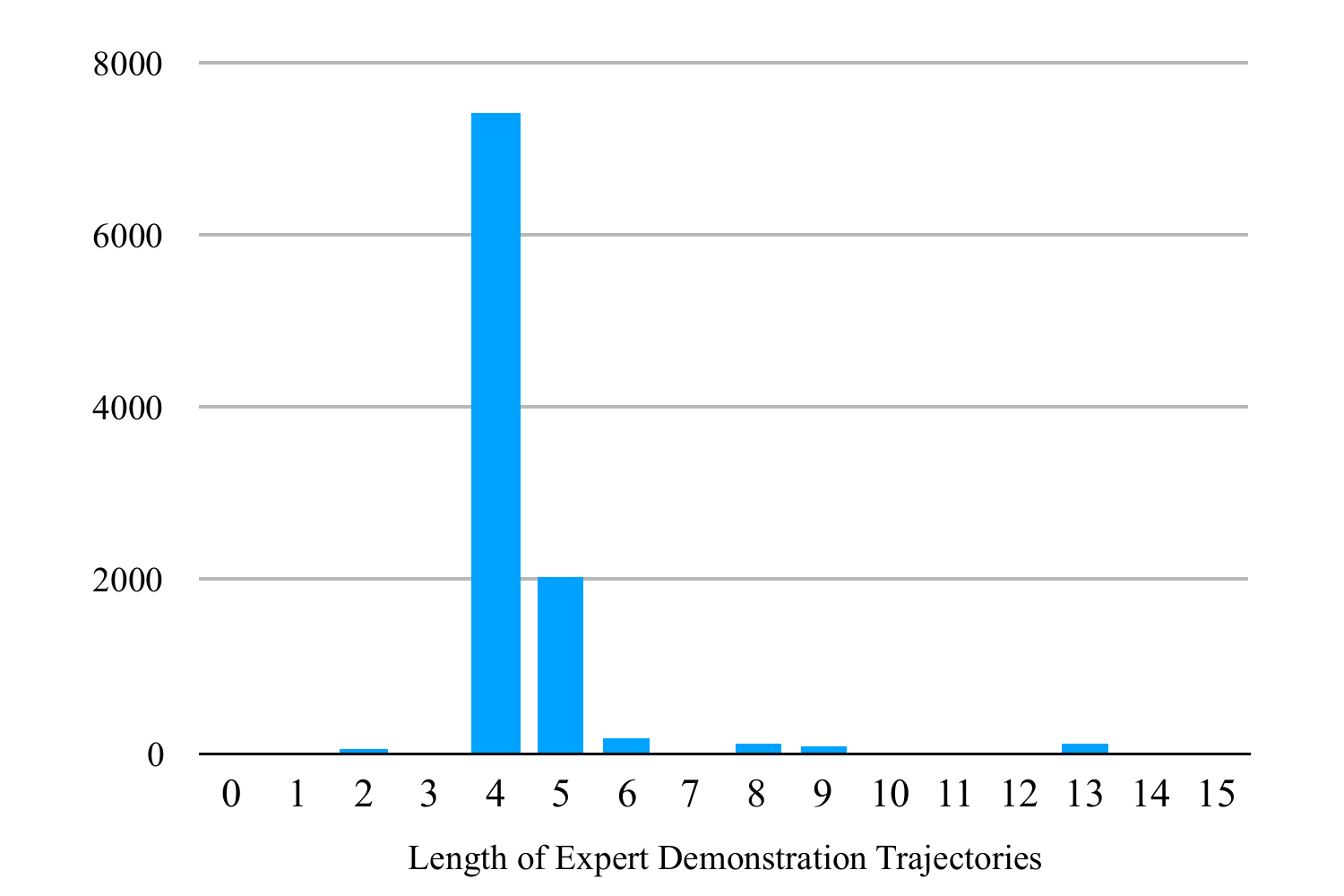}
  \caption{Lengths of trajectories given by expert demonstrations.}
  \label{fig:demo}
\end{figure}

In \fig{fig:demo} we show the distribution of lengths of expert trajectories. Most of the demonstrations have 4 or 5 steps since most subgoals follow the pick-and-place template. However, it is important to note that in the partially observable setting, the robot agent should take extra steps to search for the object being referred to, although our expert demonstration generator assumes full information of the environment.

\begin{figure}[tp]
  \centering\small
  \includegraphics[width=\textwidth]{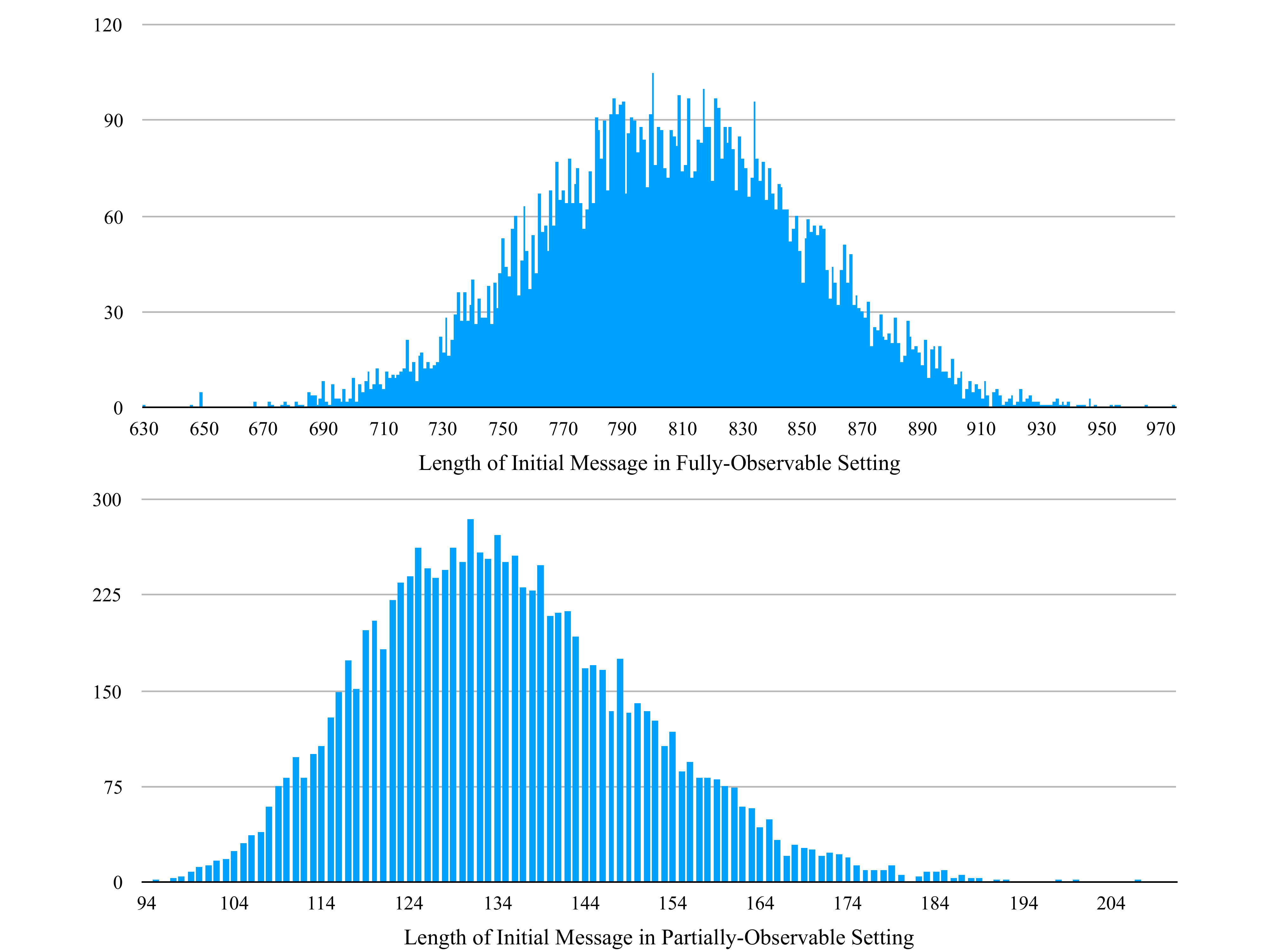}
  \vspace{-1em}
  \caption{Number of words in the initial observation in both fully- and partially-observable setting.}
  \label{fig:obs}
  \vspace{-1em}
\end{figure}

\paragraph{Textual interface.}
Presented as a text adventure game, \benchmark has a vocabulary of size 250. We present the distribution of lengths for both fully- and partially-observable setting in \fig{fig:obs}. The average length is 860 and 140 separately.

\paragraph{Data split.}
\benchmark is composed of 10,000 episodes, and can be classified in three ways. The details are shown in \tbl{tab:split}.
\textit{Hardness:} Each hardness level contains 2,500 episodes.
\textit{Quest type:}
Recall that there are 3 types of subgoals. The ``bring-me'' and ``move-to'' subgoals take the dominant part with 4,700 and 4,765 episodes separately, and the remaining 535 episodes are of the ``change-state'' subgoal.
\textit{Training split:}
We randomly split 10,000 episodes into three splits: 8,000 for training, 1,000 for validation and 1,000 for test. For each learning method, we train a series of models on the training set, choose the optimal one based on their performances on the validation set and finally, evaluate on the test set.

\begin{table}[tp]
\centering \small
\begin{tabular}{ll cccc}
\toprule
& & \textbf{Level 1} & \textbf{Level 2} & \textbf{Level 3}  & \textbf{Level 4}\\ \midrule
\multirow{3}{*}{Train}& bring-me   &1351   &711   &728   &965  \\
& move-to   &558   &1061   &1167   &1037  \\
& change-state   &78   &242   &80   &22  \\ \midrule
\multirow{3}{*}{Validation}& bring-me   &174   &89   &88   &123  \\
& move-to   &73   &117   &167   &111  \\
& change-state   &9   &31   &12   &6  \\ \midrule
\multirow{3}{*}{Test}& bring-me   &175   &100   &84   &112  \\
& move-to   &69   &122   &166   &117  \\
& change-state   &13   &27   &8   &7  \\
 \bottomrule
\end{tabular}
\vspace{0.2cm}
\caption{The 10,000 \benchmark episodes classified in 3 ways.}
\label{tab:split}
\end{table}

Version 2 dataset is composed of 116,146 episodes. The number of episodes in different hardness levels are 4008, 78488, 28343, 5307, respectively. We randomly split all episodes into three splits: 104,495 for training, 5,824 for validation, and 5,827 for test. Table xxx shows the number of episodes of different goal templates on each hardness level. %
\section{Experimental Details}
\label{sec:extended-experiments}
In this section, we explain the details of our baseline models and show the experiment results with error bars. We also present another baseline learning model and discuss potential improvements over the existing baselines.

\subsection{Model Details}

\xhdr{Seq2Seq model.}
The Seq2Seq model is based on the sequence-to-sequence model for language modeling, trained with behavior cloning on expert demonstrations~\citep{bain1995framework}. Intuitively, this model learns to predict a list of tokens (action string) given the observation tokens of historical steps. We have largely followed the model design in the ALFworld environment~\citep{ALFWorld20}.

Specifically, let $\textit{obs}_0$ be the initial observation tokens, $a_0, \textit{obs}_1, a_1 \ldots a_{k-1}, \textit{obs}_k$ be the trajectory of historical interactions. $a_i$ and $\textit{obs}_i$ are the action command and the observation at step $i$, respectively. Here, $\textit{obs}_0$ includes the welcome message and the initial ``examine'' results, \ie, all object states in the fully-observable setting, or objects at the current location of the robot in the partially-observable setting; $\textit{obs}_i$ includes the effect of command $a_i$ (for the partially-observable setting, it also includes the examine results at the current location). There is also a sequence of tokens called \textit{task\_desc} which contains human's previous actions as well as the language instruction. At each step, the input to the model includes \textit{task\_desc} and the sequence ``$\textit{obs}_0 \text{ [SEP] } a_0 \text{ [SEP] } \textit{obs}_1 \ldots a_{k-1}\text{ [SEP] } \textit{obs}_k$,'' where ``$\text{[SEP]}$'' is the separator symbol in the BERT Tokenizer~\cite{devlin-etal-2019-bert}. The model encodes the two parts of the input and fuse the encoded feature. The output of the model is $a_k$, the next action command.

\xhdr{DRRN model.}
The DRRN model is a choice-based text-game agent based on Deep Reinforcement Relevance Network (DRRN), which learns a Q function for possible state-action pairs. Concretely, the model 1) uses a GNU to encode a valid action; 2) uses separate GNUs to encode last observation, current inventory, and current examine results; 3) concatenates the above parts as the input for network; 4) outputs an estimation of Q value. Note that we include the task description information (human's past actions and the utterance) in the ``inventory'' part. The model sequentially chooses the best action based on the estimated Q values.

The network is trained by a batch of transitions tuples $(\textit{obs}, a, r, \textit{obs}', A_{\textit{valid}}(s'))$, where $\textit{obs}, \textit{obs}'$ are the observations before and after action $a$, $r$ is the reward, and $A_{\textit{valid}}(s')$ is the set of valid actions for next state $s'$. The training objective is to minimize $\|r + \gamma \max_{a'} Q(\textit{obs}', a')-Q(\textit{obs}, a)\|$.

\xhdr{Offline-DRRN model.}
Here we present an offline variation of the DRRN model. Instead of actively collecting environmental trajectories based on the current policy (with epsilon-greedy exploration), we train the Q function network with expert demonstration trajectories. Intuitively, this offline version only trains on the trajectories that can lead to positive rewards in a few steps. We show the performance of this model in the extended results below.

\subsection{Results.}
We evaluate all models on both fully-observable and partially-observable settings. We consider three evaluation metrics: 1) the average score of the model; 2) the success rate that the model achieves the goal within limited steps (40); 3) the average number of moves of successful episodes. All three scores are averaged on 1,000 episodes in the test splits.

\tbl{tab:results-fully} and \tbl{tab:results-partially} gives the experiment results for all baseline models with error bars. These results are the average values and standard deviation over three runs. Overall, the updated error bars do not change the conclusions and analysis in the main paper.

\xhdr{Analysis of the Heuristic model.}
\camera{The heuristic model leverages additional groundtruth information that all learning models do not have. Specifically, the heuristic model works on the symbolic state composed of object properties and relations. It also has a built-in planner to navigate to and manipulate objects. The only decision made by the model is to choose the proper objects to manipulate. The heuristic model performs well because the "repeating human's action" strategy is a good heuristic for using human history trajectory. To fairly compare it with the other baselines, we only allow the model to perform in a one-trial manner, \ie, the model could only guess once about the specified subgoal and then follow the plan towards it.}

\xhdr{Analysis of the Offline-DRRN model.}
Unfortunately, the offline-DRRN model also fails on \benchmark. Although the training loss decreases quickly, the model encounters a Q-value overestimation problem. Due to the generation procedure of expert demonstrations, all trajectories are successful. As a result, the model tends to attribute the success only to the last action in each trajectory, which is typically ``put-down,'' ``heat,'' ``cool,'' \etc. Empirically, these actions have high Q-values at all states. Thus, at performance time, the robot put the objects down immediately after picking up, and tends to move to fridge and microwave (in order to execute cooling and heating actions).
These issues can be potentially addressed by considering uncertainty of unseen actions in expert demonstrations (\eg, through Conservative Q-Learning~\cite{10.5555/3495724.3495824}), but are beyond the scope of this paper.

\xhdr{Reward shaping on DRRN model.}
\revise{As an extension, we perform additional reward shaping for the DRRN baseline based on expert demonstrations. Specifically, we assign positive rewards for the actions within the expert demonstrations. The $i$-th action in the sequence is assigned with extra reward $10i$. In the training phase, the robot can receive the corresponding reward if it has taken all the previous actions in the expert demonstration and then chooses the correct next action. Intuitively, this method encourages the RL agent to follow the optimal action sequence. However, training with this modified reward function does not improve the overall performance. That is, the overall success rate is still 0.00\% for all the settings. When we visualize the predicted actions of the agent, we find that in many cases the models manage to predict the first or second action correctly, but they still fail to complete the whole task. We believe that a better design in integrating imitation learning and reinforcement learning~\citep{Ross2011ARO} is needed to tackle with \benchmark tasks.}

\begin{table}[tp]
\centering \small
\begin{tabular}{l cccc}
\toprule
\multirow{2}{*}{Model} &\multicolumn{4}{c}{Fully Observable} \\ \cmidrule{2-5} 
                       & Level 1      & Level 2     & Level 3 & Level 4\\ \midrule
\multirow{3}{*}{Human}
&\mycellc{$\text{N/A}$\\$92$\\$4.4$}
&\mycellc{$\text{N/A}$\\$80$\\$4.8$}
&\mycellc{$\text{N/A}$\\$36$\\$4.3$}
&\mycellc{$\text{N/A}$\\$16$\\$4.5$}
\\ \midrule
\multirow{3}{*}{Random}
&\mycellc{$-40.0$\\$0.0$\\$\text{N/A}$}
&\mycellc{$-39.8$\\$0.1$\\$30.0$}
&\mycellc{$-40.0$\\$0.0$\\$\text{N/A}$}
&\mycellc{$-40.0$\\$0.0$\\$\text{N/A}$}
\\ \midrule
\multirow{3}{*}{Heuristic (1-trial)}   
&\mycellc{$88.9 $\\$94.9$\\$4.2$}  
&\mycellc{$-1.9 $\\$28.1$\\$4.4$}
&\mycellc{$-23.7 $\\$12.0$\\$4.3$}
&\mycellc{$ -15.9 $\\$17.8$\\$4.4$}
\\ \midrule
\multirow{3}{*}{Seq2Seq}   
&\mycellc{$-9.49\pm0.47 $\\$22.44\pm0.34$\\$4.03\pm0.01$}  
&\mycellc{$-30.74\pm0.63 $\\$6.83\pm0.46$\\$4.29\pm0.04$} 
&\mycellc{$-34.56\pm0.17 $\\$4.01\pm0.13$\\$4.29\pm0.01$} 
&\mycellc{$-32.51\pm0.00 $\\$5.51\pm0.00$\\$4.13\pm0.02$}   \\\midrule

\multirow{3}{*}{Seq2Seq+goal}  
&\mycellc{$-15.71\pm0.87$\\$17.90\pm0.64$\\$4.30\pm0.05$} 
&\mycellc{$-25.83\pm0.00$\\$10.44\pm0.00$\\$4.31\pm0.03$}   
&\mycellc{$-30.54\pm0.42$\\$6.98\pm0.31$\\$4.38\pm0.03$} 
&\mycellc{$-29.07\pm0.47$\\$8.05\pm0.34$\\$4.04\pm0.03$}  
\\\midrule

\multirow{3}{*}{Seq2Seq+subgoal}  
&\mycellc{$-4.86\pm0.22$\\$25.88\pm0.16$\\$4.20\pm0.00$}  
&\mycellc{$-22.01\pm0.00$\\$13.25\pm0.00$\\$4.20\pm0.01$} 
&\mycellc{$-21.03\pm0.00$\\$13.95\pm0.00$\\$4.03\pm0.00$}  
&\mycellc{$-19.29\pm0.00$\\$15.25\pm0.00$\\$4.22\pm0.00$}  
\\\midrule

\multirow{3}{*}{DRRN}
&\mycellc{$-39.82\pm0.18$\\$0.13\pm0.13$\\$5.00\pm0.00$}
&\mycellc{$-39.83\pm0.17$\\$0.13\pm0.13$\\$10.00\pm0.00$}
&\mycellc{$-39.83\pm0.17$\\$0.13\pm0.13$\\$9.00\pm0.00$}
&\mycellc{$-40.00\pm0.00$\\$0.00\pm0.00$\\$\text{N/A}$}\\\midrule
\multirow{3}{*}{offline-DRRN}
&\mycellc{$-40.00\pm0.00$\\$0.00\pm0.00$\\$\text{N/A}$}
&\mycellc{$-40.00\pm0.00$\\$0.00\pm0.00$\\$\text{N/A}$}
&\mycellc{$-40.00\pm0.00$\\$0.00\pm0.00$\\$\text{N/A}$}
&\mycellc{$-40.00\pm0.00$\\$0.00\pm0.00$\\$\text{N/A}$}\\
\bottomrule
\end{tabular}
\vspace{0.2cm}
\caption{Experiment results in the fully observable setting. Each model is evaluated on 4 hardness levels with 3 metrics, presented as three values in each cell: 1) the average score, 2) the success rate (\%), and 3) the average number of moves in successful episodes. The results for learning models are the mean and standard error values over three runs. }
\label{tab:results-fully}
\vspace{-1.5em}
\end{table}
\begin{table}[tp]
\centering \small
\begin{tabular}{c cccc}
\toprule
\multirow{2}{*}{Model} &\multicolumn{4}{c}{Partially Observable} \\ \cmidrule{2-5} 
                       & Level 1      & Level 2     & Level 3 & Level 4\\ \midrule
\multirow{3}{*}{Human}
&\mycellc{$\text{N/A}$\\$76$\\$5.1$}
&\mycellc{$\text{N/A}$\\$52$\\$4.8$}
&\mycellc{$\text{N/A}$\\$20$\\$5.8$}
&\mycellc{$\text{N/A}$\\$8$\\$5.0$}
\\ \midrule
\multirow{3}{*}{Random}
&\mycellc{$-40.0$\\$0.0$\\$\text{N/A}$}
&\mycellc{$-39.8$\\$0.1$\\$16.0$}
&\mycellc{$-40.0$\\$0.0$\\$\text{N/A}$}
&\mycellc{$-40.0$\\$0.0$\\$\text{N/A}$}
\\ \midrule
\multirow{3}{*}{Seq2Seq}  
&\mycellc{$-5.39\pm0.21$\\$25.49\pm0.16$\\$4.19\pm0.01$}  
&\mycellc{$-25.81\pm0.45$\\$10.44\pm0.89$\\$4.31\pm0.02$} 
&\mycellc{$-34.73\pm0.43$\\$3.88\pm0.31$\\$4.05\pm0.04$}  
&\mycellc{$-32.80\pm0.71$\\$5.30\pm0.52$\\$4.08\pm0.01$}  
\\\midrule
\multirow{3}{*}{Seq2Seq+goal}  
&\mycellc{$-11.45\pm0.00$\\$21.01\pm0.00$\\$4.13\pm0.02$} 
&\mycellc{$-28.80\pm0.67$\\$8.24\pm0.49$\\$4.05\pm0.00$}   
&\mycellc{$-30.01\pm0.00$\\$7.36\pm0.00$\\$4.32\pm0.04$} 
&\mycellc{$-32.51\pm0.71$\\$5.51\pm0.52$\\$4.04\pm0.03$}  
\\\midrule

\multirow{3}{*}{Seq2Seq+subgoal}  
&\mycellc{$-10.12\pm0.65$\\$21.99\pm0.48$\\$4.09\pm0.00$}  
&\mycellc{$-18.72\pm0.45$\\$15.66\pm0.33$\\$4.12\pm0.02$} 
&\mycellc{$-24.46\pm0.22$\\$11.44\pm0.16$\\$4.13\pm0.00$}  
&\mycellc{$-21.87\pm0.24$\\$13.35\pm0.18$\\$4.11\pm0.01$}  
\\\midrule

\multirow{3}{*}{DRRN}
&\mycellc{$-40.00\pm0.00$\\$0.00\pm0.00$\\$\text{N/A}$}
&\mycellc{$-40.00\pm0.00$\\$0.00\pm0.00$\\$\text{N/A}$}
&\mycellc{$-40.00\pm0.00$\\$0.00\pm0.00$\\$\text{N/A}$}
&\mycellc{$-40.00\pm0.00$\\$0.00\pm0.00$\\$\text{N/A}$}\\\midrule
\multirow{3}{*}{offline-DRRN}
&\mycellc{$-40.00\pm0.00$\\$0.00\pm0.00$\\$\text{N/A}$}
&\mycellc{$-40.00\pm0.00$\\$0.00\pm0.00$\\$\text{N/A}$}
&\mycellc{$-40.00\pm0.00$\\$0.00\pm0.00$\\$\text{N/A}$}
&\mycellc{$-40.00\pm0.00$\\$0.00\pm0.00$\\$\text{N/A}$}
\\
 \bottomrule
\end{tabular}
\vspace{0.2cm}
\caption{Experiment results in the partially observable setting. Each model is evaluated on 4 hardness levels with 3 metrics, presented as three values in each cell: 1) the average score, 2) the success rate (\%), and 3) the average number of moves in successful episodes. The results for learning models are the mean and standard error values over three runs. }
\label{tab:results-partially}
\vspace{-1.5em}
\end{table}

\section{Extension: Human-Robot Interaction}
\label{sec:human-interaction}
\label{sec:extension}

When the provided information is limited (\eg, in the hardness level 4), it is possible to extend the current framework to allow robot agents to ask human questions. Specifically, we can allow robots to ask questions about additional specifiers of the referred objects. As a preliminary version, we list all the valid questions in \tbl{tab:questions}. These questions query the value of a single attribute, and we can simulate the human agent response using the corresponding attribute value of the internal meaning representation $m$.  When the attribute is not specified in the meaning $m$, the human agent may return ``either is fine'' or similar answers.
\begin{table}[tp!]
\centering\small
\begin{tabular}{ll}
\toprule
        \textbf{Valid Questions} & \textbf{Example for Answers} \\ \midrule
\multirow{3}{*}{Which type do you mean?}
& I mean the apple.\\
& I just want a fruit.\\
& I just want a food.\\
\midrule
Which color do you like?
& The red one.\\
Which size do you like?
& The large one.\\
\midrule
Where is the object you want?
& On top of the table.\\
Where do you want to place it?
& Put it on the countertop.\\
\midrule
Do you want a dusty/cooked/frozen/
& Yes, I mean the dusty one.\\

sliced/toggled/soaked/open one?
& No, I mean the no dusty one.\\
\midrule
Can you say it clearly?
& I mean the sliced apple on the table.\\
\midrule
\multirow{2}{*}{(No specified answer.)}
& Not any specific type/color/\ldots\\
& Either is fine.\\

\bottomrule

\end{tabular}
\vspace{0.2cm}
\caption{All the valid questions in our textual interface and the corresponding answers. }
\vspace{-0.2cm}
\label{tab:questions}
\end{table}

Note that there should be a balance between environmental attempts and question asking. As a formalization, we can set the question cost to be a non-zero constant per specifier in the textual interface. During evaluation, the number of asked questions can also be evaluated separately.

Due to the poor performances of all learning-based models on original \benchmark setting, we do not describe this additional human-robot interaction interface in the main text. However, this is a promising extension because the real-world environments are not restricted to one single episode and it also introduces new machine learning challenges including generating helpful questions to human and learning to trade-off between question asking and environmental explorations. %

\end{document}